%% file: main.tex
\documentclass[a4paper,fleqn]{cas-dc}

\usepackage[numbers]{natbib}
\usepackage{lipsum}
\usepackage{siunitx}
\usepackage{subcaption}
\usepackage{makecell}
\usepackage[switch]{lineno}
\newcommand*{\bfrac}[2]{\genfrac{\lbrace}{\rbrace}{0pt}{}{#1}{#2}}

\definecolor{LightYellow}{RGB}{255,255,128}
\usepackage{soul}
\setul{-1.5ex}{2ex}
\setulcolor{LightYellow}

\usepackage{courier}
\usepackage{stfloats}
\usepackage[export]{adjustbox}
\usepackage{silence}
\def\tsc#1{\csdef{#1}{\textsc{\lowercase{#1}}\xspace}}
\tsc{WGM}
\tsc{QE}
\tsc{EP}
\tsc{PMS}
\tsc{BEC}
\tsc{DE}

\newcommand{\etal}{\textit{et al.}}

\begin{document}
\let\WriteBookmarks\relax
\def\floatpagepagefraction{1}
\def\textpagefraction{.001}

\shorttitle{TransformEEG: Towards Improving Model Generalizability in Deep Learning-based EEG Parkinson's Disease Detection}  

\shortauthors{F. Del Pup et~al.}  

\title [mode = title]{TransformEEG: Towards Improving Model Generalizability in Deep Learning-based EEG Parkinson's Disease Detection}  


\tnotetext[1]{
This document is the results of the research project by the European Union’s Horizon Europe research and innovation programme under Grant agreement no 101137074 - HEREDITARY.
}


\author[1,2,3]{Federico {Del Pup}}[
    type=author,
    auid=000,
    bioid=1,
    orcid=0009-0004-0698-962X
]
\cormark[1]


\ead{federico.delpup@studenti.unipd.it}


\credit{
    Conceptualization,
    Methodology,
    Software,
    Visualization,
    Writing – original draft
}

\author[1]{Riccardo {Brun}}[
    type=author,
    auid=001,
    bioid=2,
]
\ead{riccardo.brun.1@studenti.unipd.it}
\credit{
    Methodology,
    Software,
    Writing - review and editing
}

\author[1]{Filippo {Iotti}}[
    type=author,
    auid=002,
    bioid=3,
]
\ead{filippo.iotti@studenti.unipd.it,}
\credit{
    Methodology,
    Software,
    Writing - review and editing
}

\author[1]{Edoardo {Paccagnella}}[
    type=author,
    auid=003,
    bioid=4,
]
\ead{edoardo.paccagnella.1@studenti.unipd.it}
\credit{
    Software (EEGConformer),
    Writing - review and editing
}

\author[1]{Mattia {Pezzato}}[
    type=author,
    auid=004,
    bioid=5,
]
\ead{mattia.pezzato.3@studenti.unipd.it}
\credit{
    Software (Phase Swap)
    Writing - review and editing
}

\author[1]{{Sabrina} Bertozzo}[
    type=author,
    auid=005,
    bioid=6,
]
\ead{sabrina.bertozzo@studenti.unipd.it}
\credit{
    Writing - review and editing
}

\author[2,3]{Andrea {Zanola}}[
    type=author,
    auid=006,
    bioid=7,
    orcid=0000-0001-6973-8634
]
\ead{andrea.zanola@studenti.unipd.it}
\credit{
    Methodology,
    Writing - review and editing
}

\author[1,2,3]{{Louis Fabrice} Tshimanga}[
    type=author,
    auid=007,
    bioid=8,
    orcid=0009-0002-1240-4830
]
\ead{louisfabrice.tshimanga@unipd.it}
\credit{
    Methodology,
    Writing - review and editing
}

\author[4]{Henning {M\"uller}}[
    type=author,
    auid=008,
    bioid=9,
    orcid=0000-0001-6800-9878
]
\ead{henning.mueller@hevs.ch}
\credit{
    Writing - review and editing
}

\author[2,3,4]{Manfredo {Atzori}}[
    type=author,
    auid=009,
    bioid=10,
    orcid=0000-0001-5397-2063
]
\ead{manfredo.atzori@unipd.it}
\credit{
    Supervision,
    Funding acquisition,
    Project administration,
    Writing - review and editing
}

\affiliation[1]{
    organization={Department of Information Engineering, University of Padua}, 
    city={Padua},
    postcode={35131}, 
    country={Italy}
}

\affiliation[2]{
    organization={Department of Neuroscience, University of Padua}, 
    city={Padua},
    postcode={35121}, 
    country={Italy}
}

\affiliation[3]{
    organization={Padova Neuroscience Center, University of Padua}, 
    city={Padua},
    postcode={35129}, 
    country={Italy}
}

\affiliation[4]{
    organization={Information Systems Institute, University of Applied Sciences Western Switzerland (HES-SO Valais)}, 
    city={Sierre},
    postcode={3960}, 
    country={Switzerland}
}

\cortext[1]{Corresponding author}



\begin{abstract}
Electroencephalography (EEG) is establishing itself as an important, low-cost, noninvasive diagnostic tool for the early detection of Parkinson’s Disease (PD).
In this context, EEG-based Deep Learning (DL) models have shown promising results due to their ability to discover highly nonlinear patterns within the signal.
However, current state-of-the-art DL models suffer from poor generalizability caused by high inter-subject variability.
%
This high variability underscores the need for enhancing model generalizability by developing new architectures better tailored to EEG data.
%
This paper introduces TransformEEG, a hybrid Convolutional-Transformer designed for Parkinson's disease detection using EEG data.
Unlike transformer models based on the EEGNet structure, TransformEEG incorporates a depthwise convolutional tokenizer.
This tokenizer is specialized in generating tokens composed by channel-specific features, which enables more effective feature mixing within the self-attention layers of the transformer encoder.
%
To evaluate the proposed model, four public datasets comprising 290 subjects (140 PD patients, 150 healthy controls) were harmonized and aggregated.
A 10-outer, 10-inner Nested-Leave-N-Subjects-Out (N-LNSO) cross-validation was performed to provide an unbiased comparison against seven other consolidated EEG deep learning models.
%
TransformEEG achieved the highest balanced accuracy’s median (78.45\%) as well as the lowest interquartile range (6.37\%) across all the N-LNSO partitions.
When combined with data augmentation and threshold correction, median accuracy increased to 80.10\%, with an interquartile range of 5.74\%.
%
In conclusion, TransformEEG produces more consistent and less skewed results.
It demonstrates a substantial reduction in variability and more reliable PD detection using EEG data compared to the other investigated models.
\end{abstract}

%


\maketitle

\input{Ch1_Introduction}

\input{Ch2_Methods}

\input{Ch3_Results}

\input{Ch4_Discussions}

\input{Ch5_Conclusions}

\printcredits

\section*{Declaration of competing interest}
The authors declare that they have no known competing financial interests or personal relationships that could have appeared to influence the work reported in this paper.

\section*{Code and data availability}
The code used to produce both results and figures is openly available at \href{https://github.com/MedMaxLab/transformeeg}{https://github.com/MedMaxLab/trans-formeeg}.
All data that support the findings of this study are openly available within the OpenNeuro platform.

\section*{Acknowledgment}
This document is the result of the research project funded in part by the European Unions Horizon Europe research and innovation programme under Grant agreement no 101137074 - HEREDITARY, in part by the STARS@UNIPD funding program of the University of Padua, Italy, project: MEDMAX.

\bibliographystyle{elsarticle-num}
\bibliography{bibliography_abbreviated.bib}

\clearpage
\setcounter{table}{0}
\setcounter{figure}{0}
\input{supplementary}

\end{document}

%% file: Ch1_Introduction.tex
\section{Introduction}
\label{sec: intro}

Parkinson's Disease (PD) is the second most common chronic neurodegenerative disorder.
It is a progressive disease that primarily affects individuals over the age of 65, with both incidence and prevalence steadily increasing with age \cite{PD_prevalence}.
The disease results from the death of dopaminergic neurons in the substantia nigra \cite{PD_review}.
Symptoms can vary among individuals and include motor symptoms, such as tremors, bradykinesia (slowness of voluntary movement initiation), akinesia (absence of normal unconscious movements), and hypokinesia (reduction in movement amplitude), as well as non-motor symptoms like sleep behavior disorders, constipation, and anxiety \cite{PD_motor, PD_nonmotor}.
PD symptoms can significantly reduce patients' quality of life, especially since no disease-modifying pharmacologic treatments are currently available \cite{pd_treatment}.
Therefore, early diagnosis of PD is of clinical significance, as it can help improve patients' quality of life by potentially slowing disease progression.

Electroencephalography (EEG) is establishing itself as an important, low-cost, noninvasive diagnostic tool for the early detection of PD.
This success stems from the potential use of quantitative EEG measures as biomarkers of disease severity and progression \cite{pd_eeg2}.
In \cite{pd_eeg1}, the analysis of 36 different studies confirmed that both global and domain-specific cognitive impairments correlated with EEG ``slowing" \cite{pd_slow}.
EEG slowing is associated with alterations of the normal oscillatory brain activity, characterized by decreased spectral power in alpha and beta bands and increased spectral power in delta and theta bands.

To identify differences between healthy and PD individuals in EEG recordings, several automatic classification approaches have been investigated \cite{pd_ml}.
Among these, methods based on deep learning (DL) have shown promising results due to the ability of neural networks to discover highly nonlinear patterns embedded within the signal \cite{pd_dl_survey}.
Various architectures have been explored, including Convolutional Neural Networks (CNN) \cite{pdcnnet}, Recurrent Neural Networks (RNN) \cite{pd_rnn}, and, more recently, transformer models \cite{dicenet}.

However, current state-of-the-art EEG deep learning models suffer from poor generalizability, underscored by their high sensitivity to variation in the experimental setting.
In \cite{DelPup2024b}, it was found that variations in the preprocessing pipeline can lead to significant fluctuations in model accuracy.
Focusing on the PD detection results reported in this study, when data are preprocessed with only a minimal filtering, median accuracy reaches 66\%.
Introducing independent components rejection \cite{ICLabel} increases the median accuracy to 75\%, but adding more advanced artifact removal techniques (e.g., artifact subspace reconstruction \cite{asr}) causes the metric to drop back down to 67\%.
These results show that certain EEG artifact can improve the accuracy of deep learning models but can also alter the quality of the learned features, ultimately reducing their generalizability.

Similarly, the way EEG data are partitioned can greatly overestimate the model's performance and generalizability.
In \cite{NLOSO}, it was demonstrated that EEG deep learning models can reach almost perfect accuracy (with little variability) if segments from the same EEG recordings are assigned to both the training and test sets.
However, when a cross-subject analysis is performed, accuracy drops of over 20\%, accompanied by an alarming increase in result variability. 
This variability suggests that inherent characteristics of the EEG signal, such as biometric features and potential correlations across consecutive segments, can be exploited by the model to solve the task at hand without necessarily learning features that are truly representative of the underlying pathology.

Recently proposed EEG-DL models incorporate attention layers to better identify and capture long-range patterns within the signal \cite{eegconformer, atcnet}.
However, these models often use EEGNet-like convolutional encoders to generate the sequence of tokens that are fed into the transformer blocks \cite{eegnet}, which limits the capabilities of the attention layers.
While the addition of transformer layers on top of an EEGNet-like convolutional tokenizer can improve model performance, it does not address the generalizability issue reported in \cite{NLOSO} using the same family of models, which might stem from how tokens are generated.

When evaluated using unbiased cross-validation methods such as the Nested-Leave-N-Subjects-Out (N-LNSO), EEG deep learning models demonstrate lower generalizability, underscored by lower performances with higher variability \cite{NLOSO}.
This variability not only makes it difficult to compare models fairly, but it also underscores the need for enhancing model generalizability by developing new architectures that are better tailored to EEG data.
Enhancing model generalizability is crucial for increasing the reliability of EEG deep learning systems and enabling their application in clinical scenarios.

\textbf{Contributions:} This study introduces TransformEEG, a novel hybrid Convolutional-Transformer model designed for PD detection using EEG data.
Unlike transformer models based on the EEGNet structure, TransformEEG incorporates a carefully designed depthwise convolutional tokenizer.
This tokenizer specializes in generating tokens that are composed by channel-specific features, enabling more effective feature mixing within the transformer encoder.
TransformEEG is evaluated on Parkinson's disease classification against seven other consolidated EEG deep learning models, showing better results in terms of balanced accuracy's median and interquartile range.
To provide reliable performance estimates, the evaluation is conducted on four publicly available datasets comprising 290 subjects (140 PD patients, 150 healthy controls), using a 10-outer, 10-inner Nested-Leave-N-Subjects-Out cross-validation scheme.

\textbf{Paper structure:} the outline of this paper is as follows.
Section \ref{sec: methods} describes in detail the experimental setting.
Section \ref{sec: results} presents the comparative analysis between TransformEEG and seven other EEG deep learning models.
Section \ref{sec: discussion} critically discusses the results, highlighting potential limitations and future directions.
Finally, a conclusion is drawn in Section \ref{sec: conclusion}.

%% file: Ch2_Methods.tex
\section{Methods}
\label{sec: methods}

This section outlines key methodological aspects of the study.
First, it provides a concise description of the selected datasets and their preprocessing steps.
Next, it introduces the proposed TransformEEG architecture.
Finally, it details additional training information, including data partitioning, data augmentation, training hyperparameters, and model evaluation strategies.
Together with the openly available source code repository and supplementary materials, this section provides all the features listed in \cite{DelPup2024a} to enhance the reproducibility of this study.

\subsection{Dataset selection}
\label{subsec: data sel}

The analysis was conducted on four open-source datasets.
These datasets are briefly described in the following subsections, with additional details provided in Table \ref{tab: datasets}.
Each dataset is openly available within OpenNeuro\footnote{[Online] Available: \href{https://openneuro.org/}{https://openneuro.org/}} \cite{OpenNeuro}, an established open platform for sharing neuroimaging data organized in BIDS format \cite{EEGBIDS}.

\subsubsection{Dataset 1: ds004148 - EEG test-retest}
This dataset \cite{ds004148} comprises resting-state (both eyes-open and eyes-closed) and cognitive state recordings from 60 healthy subjects, with an average age of $20.0\!\pm\!1.9$ years.
All subjects participated in three recording sessions, during which both resting-state and cognitive tasks were performed.
To prevent unbalancing the sample distribution across different datasets, only the resting-state recordings from session one were considered, as done in \cite{DelPup2024b}.
The selected resting-state recordings have a fixed duration of exactly 300 seconds.

\subsubsection{Dataset 2: ds002778 - UC San Diego}
This dataset \cite{ds002778} includes resting-state, eyes-open EEG recordings from 15 individuals with Parkinson's disease and 16 age-matched healthy controls.
The average age of the two groups is $63.3 \pm 8.2$ years for the Parkinson's patients and $63.5 \pm 9.7$ years for the healthy controls.
Healthy subjects participated in only one session, while Parkinson's individuals underwent two sessions: the first, recorded after they discontinued medication for at least 12 hours before the session (\textit{ses-off}); the second, recorded while they were under medication (\textit{ses-on}).
For the subsequent analysis, only the off-medication recordings were included.
The selected recordings have an average duration of $195.7 \pm 18.8$ seconds.

\subsubsection{Dataset 3: ds003490 - EEG 3-Stim}
This dataset \cite{ds003490} comprises resting-state EEG recordings (both eyes-open and eyes-closed) and auditory oddball tasks from 25 individuals with Parkinson's disease and 25 age-matched healthy controls.
The average age of the two groups is $69.7 \pm 8.7$ years for the Parkinson's patients and $69.3\pm 9.6$ years for the healthy controls.
Healthy controls participated in a single recording session, while Parkinson's individuals underwent acquisitions both off-medication (at least 15 hours) and on-medication.
For the subsequent analysis, only the off-medication recordings were included.
The selected records have an average duration of $595.9 \pm 74.0$ seconds.

\subsubsection{Dataset 4: ds004584 - EEG PD}
This dataset \cite{ds004584} comprises resting-state, eyes-open EEG recordings from 100 Parkinson's patients and 49 age-matched healthy controls.
The average age of the two groups is respectively $68.5 \pm 8.1$ years for the Parkinson's patients and $70.9\pm 7.6$ years for the healthy controls.
All subjects underwent a single recording session, with an average duration of $144.4 \pm 46.0$ seconds.

\begin{table}[!th]
\centering
\caption{Dataset description.}
\begin{tabular}{cccccc}
\toprule
\makecell{dataset\\ID} & \makecell{original\\reference} & \makecell{\#\\
Chan} & \makecell{$f_s$\\$\text{[Hz]}$} & \makecell{\#\\Subj} & \makecell{\#\\Samples$^*$}
\\ \midrule
ds004148 & FCZ     & 64 & 500 & 60  & \num{2880} \\
ds002778 & CMS/DRL & 41 & 512 & 31  & \num{456}  \\
ds003490 & CPZ     & 64 & 500 & 50  & \num{2408} \\
ds004584 & PZ      & 63 & 500 & 149 & \num{1775} \\
\midrule
Total    &         &    &     & 290 & \num{7519} \\
\bottomrule
\multicolumn{6}{l}{*Assuming EEG windows of 16s and 25\% of overlap}
\end{tabular}
\label{tab: datasets}
\end{table}

\subsection{Data preprocessing}
\label{subsec: prepro}

All the selected recordings were preprocessed and harmonized using an automatic pipeline implemented in \textit{BIDSAlign}\footnote{https://github.com/MedMaxLab/BIDSAlign} \cite{BIDSAlign} (v1.0.0).
This pipeline is based on the findings from a recent analysis on the role of preprocessing in EEG deep learning applications \cite{DelPup2024b}, which showed that adding more intensive steps, such as noisy channel removal and window correction via artifact subspace reconstruction (ASR) \cite{asr}, does not improve model performance.
The final standardization, downsampling, and windows extraction steps were performed online within the Python environment during data loading for model training.
The preprocessing steps are detailed below in their execution order.

\WarningsOff
\begin{enumerate}[$\bullet$]
    \item \textit{Non-EEG channels removal}: other biosignals (e.g., ECG, EOG) included as extra channels were removed.
    \item \textit{Time segments removal}: the first and last 8 seconds of each recording were removed to exclude potential divergences in the signal. This step limits the removal of brain activity by enhancing the quality of the subsequent independent component rejection step.
    \item \textit{DC component removal}: The direct current (DC) voltage was subtracted from each EEG channel. This operation is equivalent to removing the mean from each EEG channel.
    \item \textit{Resampling}: EEG recordings were resampled to 250 Hz to reduce the computational burden of the subsequent preprocessing operations.
    \item \textit{Filtering}: EEG signals were filtered with a passband Hamming windowed sinc FIR filter with passband edges of respectively 1 Hz and 45 Hz.
    \item \textit{Automatic independent components rejection}: Independent Components (ICs) were extracted using the \textit{runica} algorithm {\cite{infomaxica}}, with no limit on the number of components.
    These components were automatically rejected using IClabel {\cite{ICLabel}} by applying the following rejection thresholds: [90\%, 100\%] confidence for components considered noisy (e.g., muscle, eye, heart, line noise, channel noise) and [0\%, 10\%] confidence for components representing brain activity, following the approach in \cite{DelPup2024b}.
    \item \textit{Re-referencing}: EEG recordings were re-referenced to the common average.
    \item \textit{Channel selection}: A subset of 32 EEG channels, common to all four selected datasets, was extracted and used for subsequent analysis.
    \item \textit{Downsampling}: EEG signals were further downsampled to 125 Hz to improve computation and reduce the GPU's memory occupation during model training. As explained in \cite{DelPup2024b}, direct resampling from 250 Hz to 125 Hz was avoided to preserve the quality of the automatic independent component rejection process.
    \item \textit{Standardization}: the $z$-score operator was applied along the EEG channel dimension. This step transforms each EEG channel into a signal with mean $\mu\text{=0}$ and standard deviation $\sigma\text{=1}$.
    \item \textit{Windows extraction}: EEG data were partitioned into 16 second windows with 25\% of overlap to increase the number of samples.
\end{enumerate}
\WarningsOn

Preprocessed data from all the four datasets were combined to construct a binary classification task that aims to distinguish between Parkinson's and healthy subjects.

\subsection{Model Architecture}
\label{subsec: transformeeg}

\begin{figure*}[!t]
    \centering
    \includegraphics[width=\textwidth]{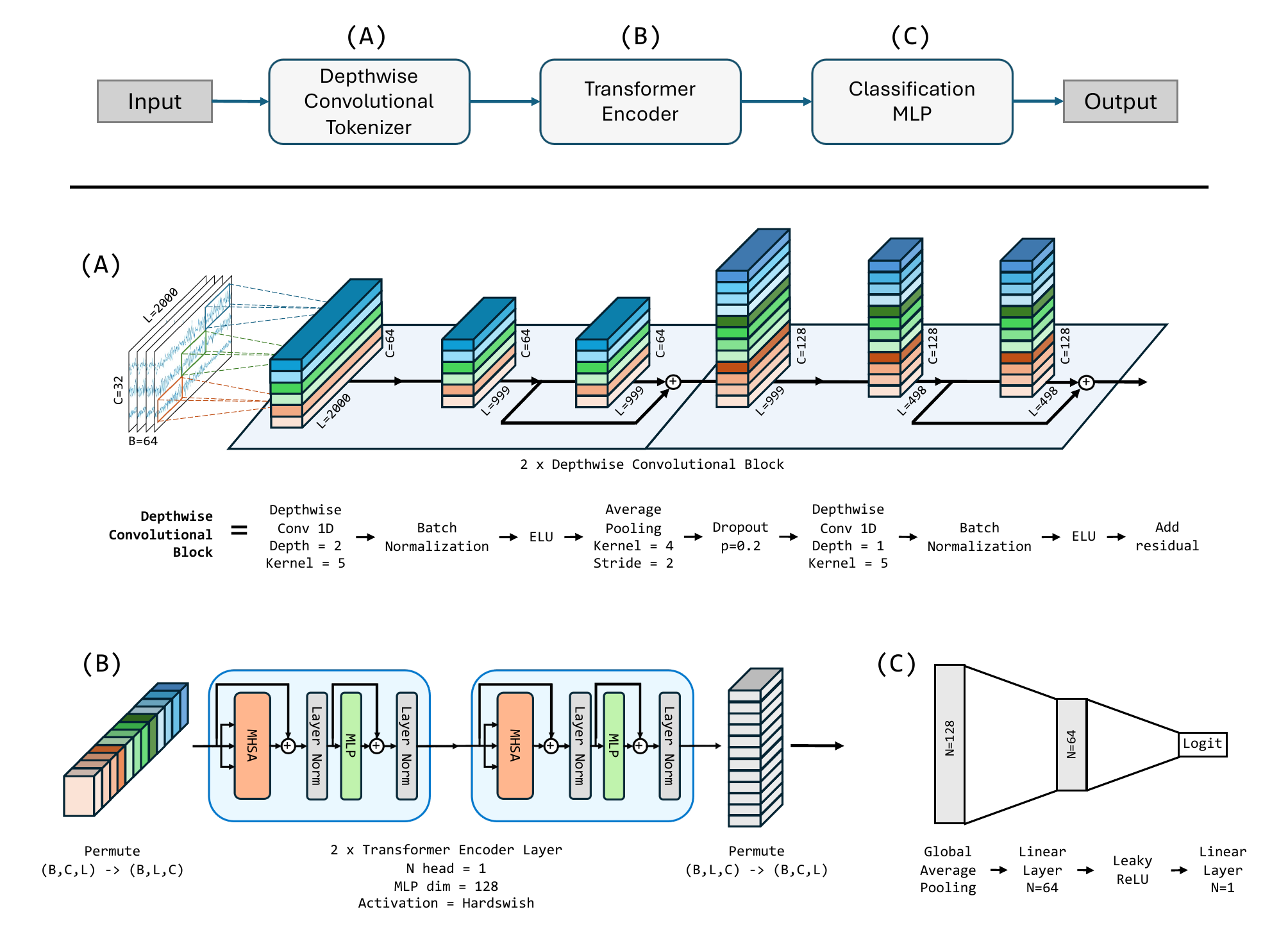}
    \caption{
    Schematic representation of the TransformEEG architecture.
    TransformEEG consists of three modules: a depthwise convolutional tokenizer (A), a transformer encoder (B), and a classification MLP (C).
    The depthwise convolutional tokenizer creates EEG tokens describing local time portions of the input window with channel-specific features.
    The transformer encoder recombines the tokens with the self-attention mechanism.
    The classification MLP outputs class predictions.
    }
    \label{fig: transformeeg_model}
\end{figure*}

TransformEEG is a hybrid convolutional-transformer architecture composed of three modules, each stacking a series of neural blocks. 
The first module is the depthwise convolutional tokenizer, which creates EEG tokens describing local portions of the input window with channel-specific features.
The second module is the transformer encoder, which recombines the generated tokens using the self-attention mechanism.
The third module is the classification MLP, which produces class predictions based on the output of the transformer encoder.
Figure \ref{fig: transformeeg_model} illustrates the structure of TransformEEG.
Table 12 in Section A.8 of the Supplementary Materials included a detailed description of the input and output dimensions and the number of parameters for each layer.
A PyTorch \cite{pytorch} implementation of the model is available in the open-source code repository\footnote{https://github.com/MedMaxLab/transformeeg}.
A description of each module is provided below.

\subsubsection{Depthwise Convolutional Tokenizer}

The depthwise convolutional tokenizer is a module designed to transform an input EEG window into a sequence of tokens. 
This module consists of two stacked depthwise convolutional blocks, which extract features based on local temporal patterns within individual EEG channels.
The depthwise convolutional block performs the following operations:

\begin{enumerate}
    \item Depthwise 1D convolution with a depth multiplier of 2, which doubles the feature dimension.
    \item Batch normalization, followed by ELU activation.
    \item Average Pooling with stride 2, which halves the sequence length.
    \item Dropout ($p=0.2$), which is applied during training for regularization.
    \item Depthwise 1D convolution with a depth multiplier of 1.
    \item Batch normalization, followed by ELU activation.
    \item Residual connection addition (from the output after dropout), to improve gradient stability.
\end{enumerate}

Depthwise convolutions ensure that, first, each feature is associated with a unique input EEG channel (even after multiple blocks are applied sequentially) and, second, that the number of parameters remains low.
Maintaining a small parameter count helps prevent overparameterization and reduces overfitting tendencies, which can be particularly detrimental in EEG-based pathology classification tasks, as discussed in \cite{sample_leakage_2} and in \cite{xeegnet}.
Average pooling halves the sequence length, thereby reducing memory usage and computational load.
As a result, the combination of depthwise convolutions and pooling produces tokens that represent local portions (consecutive time steps) of the input EEG window, effectively capturing channel-specific features.

Denoting the following variables:

\WarningsOff
\begin{enumerate}[\textbullet]
\item $B$ as the batch size,
\item $C$ as the number of input EEG channels,
\item $L$ as the window length,
\item $D$ as the depth multiplier of the depthwise convolutional layer,
\item $S$ and $K$ as the stride and kernel size of the average pooling layer, respectively,
\end{enumerate}
\WarningsOn

\noindent The combination of two depthwise convolutional blocks produces a new multidimensional array with the following dimensions:

\begin{equation}
\begin{split}
    (B,C,L) &\rightarrow \left(B,C\times D, \left\lfloor \frac{L+(S-K)(1+S)}{S^2} \right\rfloor \right)  
\end{split}
\end{equation}

\noindent Therefore, with input dimensions of (64, 32, 2000), the output dimensions are (64, 128, 498).

The TransformEEG's tokenizer module diverges from typical EEGNet-like convolutional encoders used in attention-based architectures such as EEGConformer \cite{eegconformer} or ATCNet \cite{atcnet}.
EEGNet-like encoders treat the input window as a single-channel pseudo image, processed with 2D convolutional layers that operate along either the temporal or spatial dimensions.
Consequently, spatial convolutions extract features that are linear combinations of the original EEG channels.
Additionally, since no padding is used, this process conveniently removes the EEG channel dimension and prepares the input for the transformer encoder (see Supplementary Materials, Section A.8).
However, as a result, attention layers are forced to perform linear projections on features that may not optimally represent the original channels.
This redundancy can negatively impact the effectiveness of feature mixing within the transformer encoder, as such mixing heavily depends on the quality of channel recombination performed by the spatial layers.

In contrast, each module of TransformEEG has a clear and complementary role.
The convolutional tokenizer processes EEG segments as multi-channel time-series data.
Depthwise convolutions apply a separate filter to each input EEG channel independently, aiming to extract channel-specific features based on local temporal patterns.
No channel recombination occurs at this stage, ensuring that the extracted features better reflect the properties of each individual channel.
This design potentially enhances the generalizability of the model because it allows for improved feature mixing across channels within the transformer encoder.
In summary, the complementary roles of convolutional and attention layers enhance the network's ability to capture both local and global patterns within the EEG window, leading to more consistent performance, as discussed in Section \ref{sec: results}.

\subsubsection{Transformer Module}

The sequence of tokens generated by the depthwise convolutional tokenizer is given to the transformer module (Figure \ref{fig: transformeeg_model}, Panel B), which consists of a series of transformer encoder layers.
Each transformer encoder layer processes the input EEG tokens using a self-attention mechanism and applies a subsequent non-linear transformation with a multi-layer perceptron (one hidden layer with size equals to the embedding dimension, 128).
Layer normalization is applied after each skip connection, following the approach described in \cite{attention}.
Permutation operations are included for compatibility with PyTorch library objects.
In contrast with traditional implementations, no positional embeddings are added to the sequence of tokens, nor is a special class token concatenated to the output of the convolutional tokenizer \cite{bert}. 
An ablation analysis, presented in Section A.5.1 of the Supplementary Materials, demonstrates that including one or both of these elements does not improve the performance or generalizability of TransformEEG.
The number of heads in the self-attention layers is set to one.
As discussed in Section A.5.2 of the Supplementary Materials, increasing the number of heads does not improve the model's performance.
While using more attention heads allows TransformEEG to capture multiple types of relationships within the input data simultaneously, it also increases the chances that the model will overfit the training set, especially if the number of subjects (samples) is low.

\subsubsection{Classification MLP}

The output of the transformer encoder is fed into the final classification module to generate predictions.
Since there is no class token, global average pooling is initially performed to produce an embedding vector.
This embedding vector is then fed into a multi-layer perceptron (MLP) comprising one hidden layer.
The hidden layer halves the embedding dimension and applies a Leaky ReLU activation function (negative slope of 0.01) as non-linearity. 
The MLP's output is subsequently passed through a sigmoid function to produce a probability indicating whether the sample originates from an EEG of a patient with Parkinson's disease.
Subsections \ref{subsec: baseline} and \ref{subsec: DA effect} present the results using a default probability threshold of 0.5 to classify samples as positive or negative, while subsection \ref{subsec: th effect} presents the results when correction procedures are applied to adjust this threshold.

\subsection{Implementation details}
\label{subsec: implement}
Models were trained using \textit{SelfEEG}\footnote{https://github.com/MedMaxLab/selfEEG} \cite{DelPup2024c} (v0.2.0) and figures were generated with \textit{Seaborn} \cite{seaborn}.
Experiments were conducted on an NVIDIA A30 GPU device with CUDA 12.2.
Further implementation details are provided in the supplementary materials and in the openly available source code.

\subsubsection{Data Partition and Model Evaluation}
\label{subsubsec: data partition}

After dividing the data into 16-second windows with 25\% overlap, the datasets were partitioned and evaluated using the Nested-Leave-N-Subjects-Out (N-LNSO) cross-validation (CV) method described in \cite{DelPup2024b}.
This method involves concatenating two Leave-N-Subjects-Out CV procedures to generate a set of unique train-validation-test splits.
For each split, models are trained on the training set, monitored on the validation set, and evaluated on the test set.
This study uses ten outer folds and ten inner folds, for a total of 100 splits per N-LNSO procedure.
Additional details are provided in Section A.1 of the Supplementary Materials.

Seven established deep learning models were considered for the evaluation.
The list is composed by xEEGNet \cite{xeegnet}, EEGNet \cite{eegnet}, ShallowNet \cite{shallow}, ATCNet \cite{atcnet}, EEGConformer \cite{eegconformer}, DeepConvNet \cite{shallow}, and EEGResNet \cite{resnet2}.
These models encompass different architectural type, such as convolutional (e.g., EEGNet, ShallowNet) and attention-based models (e.g., EEGConformer, ATCNet). 
The number of learnable parameters spans from few hundreds (xEEGNet, 245 parameters) to more than a million (EEGResNet, \num{1337665} parameters).
This diversity enriches the analysis and enables a comprehensive comparison with state-of-the-art models.

Models were evaluated using the balanced accuracy, defined as the macro average of recall scores per class \cite{metricsML}.
Results based on additional metrics such as F1-Score and Cohen's kappa are reported in Section A.2 of the Supplementary Material.
The results from all splits are aggregated to compare TransformEEG with the other selected EEG deep learning models.
Comparison metrics include the median value, interquartile range (IQR), and the $[1^{st}\!-99^{th}]$ percentile range.

\subsubsection{Data Augmentation}
\label{subsubsec: data augmentation}

To enable a fair comparison of the considered architectures, data augmentation effects were assessed by selecting the optimal configuration for each model.
A set of 10 different data augmentations was selected for the comparison presented in subsection \ref{subsec: DA effect}, based on dedicated analyses performed on EEG signals \cite{data_aug1, data_aug2, data_aug3, phaseswap}.
Each augmentation was investigated individually and in combination with another, resulting in a total of 100 possible data augmentation configurations.
For each of these configurations, a N-LNSO cross-validation was performed on the architectures listed in the subsection \ref{subsubsec: data partition}.
The median and interquartile range of balanced accuracy were compared to identify the most effective augmentation combination for each model.

Determining the optimal data augmentation is a challenging process, as both the median (central measure) and the interquartile range (variability) are crucial in the comparison.
In addition, these two values evolve on different scales, increasing the challenges in weighting their improvement over the reference baseline.
In order to identify a consistent and objective measure, which keeps into account the relative improvements of both the baseline median and interquartile range, this work proposes the Augmentation Relative Improvement Score (ARIS).
ARIS is defined as follows

\begin{equation}
    \textit{ARIS} = 
    \begin{cases}
        \begin{aligned}
        &0 \quad\text{if } (M_b>M)  \lor (\text{IQR}_b<\text{IQR}) \\
         &\frac{M_b-M}{M_b} \times \frac{\text{IQR}-\text{IQR}_b}{\text{IQR}_b}  \quad\text{otherwise}            
        \end{aligned}
    \end{cases}
\end{equation}
where:
\WarningsOff
\begin{enumerate}[-]
    \item $M_b$ is the median baseline without data augmentation.
    \item $M$ is the current median value with data augmentation.
    \item $\text{IQR}_b$ is the baseline interquartile range.
    \item $\text{IQR}$ is the current interquartile range.
\end{enumerate}
\WarningsOn

This scores ensures that data augmentations that do not improve both the median and interquartile range of the balanced accuracy are discarded.
Additionally, it ensures improvements are properly weighted by the scale of the baseline value.

A description of each data augmentation, along with the used set of parameters (detailed in Table \ref{tab: da hyper}), is provided below.
Their Python implementation is available within the \textit{selfEEG} source code, which is openly accessible on GitHub.
An analytical formulation is also provided to clarify their description. 
The input EEG window is denoted as a matrix $\text{X}\in\mathbb{R}^{C\times N}$, where C is the number of channels and N is the number of time steps.
The value of $N$ is calculated as $N = f_s \times L$, with $f_s$ being the sampling frequency in Hertz (125 Hz) and $L$ the sequence length in seconds (16s).
The augmented version of $\text{X}$ is denoted as $\tilde{\text{X}}~\in~\mathbb{R}^{C\times N}$.
Additionally, $x_{c,t}$ is used to indicate the value of channel $c$ at the time step $t$ of an EEG sample $\text{X}$.

\renewcommand{\arraystretch}{1.4}
\begin{table*}[h]
\caption{Data augmentation hyperparameters: hyperparameter values are randomly selected at each call to increase sample variability.}
\label{tab: da hyper}
\begin{tabular}{ccc}
\toprule
Data Augmentation & Hyperparameter Name & Hyperparameter Value \\
\midrule
    \multicolumn{1}{c|}{Sign Flip} & 
    \multicolumn{1}{c|}{None} &
    \multicolumn{1}{c}{None}
    \\ \hline
    \multicolumn{1}{c|}{Time Reverse} & 
    \multicolumn{1}{c|}{None} &
    \multicolumn{1}{c}{None} 
    \\ \hline
    \multicolumn{1}{c|}{\multirow{2}{*}{Band Noise}} & 
    \multicolumn{1}{c|}{Bandwidth} &
    \multicolumn{1}{c}{\makecell{$\text{band}\in \bfrac{\text{delta, theta, alpha}}{\text{beta, gamma low}}$}}
    \\ \cline{2-3}
    \multicolumn{1}{c|}{} &
    \multicolumn{1}{c|}{$\sigma$} &
    \multicolumn{1}{c}{$\sigma\in[0.8, 0.9]$}  \\ \hline
    \multicolumn{1}{c|}{Signal Drift} &
    \multicolumn{1}{c|}{Slope} &
    \multicolumn{1}{c}{\makecell{$m/(125 \times 16)$ \\$m \in \{\pm0.5, \pm0.4, \pm0.3\}$}} 
    \\ \hline
    \multicolumn{1}{c|}{SNR Scaling} & 
    \multicolumn{1}{c|}{SNR} &
    \multicolumn{1}{c}{$\text{SNR} \in[8, 10]$}  \\ \hline
    \multicolumn{1}{c|}{Channel Dropout} &
    \multicolumn{1}{c|}{Channels to drop}&
    \multicolumn{1}{c}{$n \in [4, 16],\space n\in \mathbb{Z}$}
    \\ \hline
    \multicolumn{1}{c|}{\multirow{2}{*}{Masking}} &
    \multicolumn{1}{c|}{Masked portions} &
    \multicolumn{1}{c}{$k \in \{3, 4, 5\}$} 
    \\ \cline{2-3}
    \multicolumn{1}{c|}{} &
    \multicolumn{1}{c|}{Masking ratio} &
    \multicolumn{1}{c}{$p\in[0.2, 0.35]$}  \\ \hline
    \multicolumn{1}{c|}{\multirow{3}{*}{Signal Warp}} &
    \multicolumn{1}{c|}{Number of segments} &
    \multicolumn{1}{c}{$n\in \{6, 7, 8\}$}
    \\ \cline{2-3} 
    \multicolumn{1}{c|}{} &
    \multicolumn{1}{c|}{Stretch strength} &
    \multicolumn{1}{c}{$k_{st}\in[1.25, 1.50]$}
    \\ \cline{2-3}
    \multicolumn{1}{c|}{} &
    \multicolumn{1}{c|}{Squeeze strength} &
    \multicolumn{1}{c}{$k_{sq}=1$}  \\ \hline
    \multicolumn{1}{c|}{Phase Randomizer} & 
    \multicolumn{1}{c|}{perturbation strength} &
    \multicolumn{1}{c}{$s=0.9$}
    \\ \hline
    \multicolumn{1}{c|}{Phase Swap} &
    \multicolumn{1}{c|}{None} &
    \multicolumn{1}{c}{None}
    \\
\bottomrule
\end{tabular}
\end{table*}
\renewcommand{\arraystretch}{1}

\WarningsOff
\begin{enumerate}[\textbullet] 
    \item \textbf{Time Reverse}: This transformation flips the signal horizontally by applying the following operation: 
    \begin{equation}
        \tilde{x}_{c,t} = x_{c,L-t}
    \end{equation}
    \item \textbf{Sign Flip}: The signal is flipped vertically. This data augmentation simulates an inversion of polarity in the electrodes and it is described by the transformation
    \begin{equation}
        \tilde{\text{X}} = -\text{X}
    \end{equation}
    \item \textbf{Band Noise}: This augmentation adds random noise filtered within specific EEG bandwidths. Given the impulse response $h(t)$ of a bandpass filter that preserves a particular EEG bandwidth, the augmented EEG sample can be defined as:
    \begin{equation}
        \tilde{x}_{c,t} = x_{c,t} + (\varepsilon \ast h)(t)
    \end{equation}
    where $\varepsilon(t)$ is Gaussian noise with variance $\sigma^2$, an augmentation hyperparameter, and $\ast$ denotes the convolution operation.
    \item \textbf{Signal drift}: The signal is drifted with either a positive or negative slope by applying the transformation:
    \begin{equation}
        \tilde{x}_{c,t} = x_{c,t} + mt
    \end{equation}
    where $m$ is an augmentation hyperparameter that determines the slope of the drift.
    \item \textbf{SNR Scaling}: This augmentation adds random noise to an EEG sample, scaled to reach a specified signal-to-noise ratio (SNR).
    Given a target SNR, the augmented EEG sample can be defined as:
    \begin{equation}
        \tilde{x}_{c,t} = x_{c,t} + k\varepsilon_t
    \end{equation}
    where 
    \begin{enumerate}[-] 
        \item $\varepsilon_t \sim \mathcal{N}(0, 1)$ is Gaussian noise with unitary variance
        \item $k$ is a scaling factor calculated as
    \end{enumerate} 
    \begin{equation}
        k = 10^{(-\frac{\text{SNR}}{20})} \sqrt{\frac{1}{NC}\sum_{c}\sum_{t}x_{c,t}^2}
    \end{equation}
    \item \textbf{Channel dropout}: This augmentation sets a predefined number of EEG channels to zero. Specifically, given a dropout percentage $p$ and a bijective function $\sigma : S \rightarrow S$ that defines a random permutation of the elements of a vector, the augmented EEG sample is computed as
    \begin{equation}
        \tilde{\text{X}} = \text{X} \odot (\mathbf{s}_C \mathbf{1}_N^T)
    \end{equation}
    where
    \begin{equation}
        \mathbf{s}_C = \sigma(\mathbf{0}_{\lfloor Cp \rfloor} \oplus \mathbf{1}_{C-\lfloor Cp \rfloor} )
    \end{equation}
    In this context:
    \begin{enumerate}[-]
        \item $\odot$ is the Hadamard product, representing the element-wise product of two matrices.
        \item $\mathbf{s}_C$ is a random column vector composed of zeros and ones, indicating which channels are set to zero.
        \item The permutation $\sigma$ shuffles the elements of this vector.
        \item $\oplus$ is the operator describing the concatenation of two vectors.
    \end{enumerate}
    \item \textbf{Signal masking}: This augmentation sets portions of the signals to zero. Specifically, given a masking percentage $p$ and a predefined number of masking blocks $k$, the augmented signal can be defined as:
    \begin{equation}
        \tilde{\text{X}} = \text{X} \odot ( \mathbf{1}_C \mathbf{m}^T)
    \end{equation}
    where:
    \begin{enumerate}[-]
        \item $\mathbf{1}_C$ is a column vector of ones with length equal to the number of channels $C$,
        \item $\mathbf{m}$ is a column mask vector of length equal to the signal length, constructed as a concatenation of blocks of ones and zeros, i.e.,
        \begin{equation}
            \mathbf{m} = \mathbf{1}_{b_1} \oplus \mathbf{0}_{b_2} \oplus \mathbf{1}_{b_3} \oplus \mathbf{0}_{b_4} \oplus \dots \oplus \mathbf{0}_{b_{2k}} \oplus \mathbf{1}_{b_{2k+1}}
        \end{equation}
        with each $b_i$ representing the length of the corresponding block. The total length of $\mathbf{m}$ is:
        \begin{equation*}
            N = \sum_{i=1}^{2k+1} b_i
        \end{equation*}
        and the total number of masked samples, corresponding to the sum of the zero blocks, is
        \begin{equation*}
            N\!p = \sum_{\substack{i=2,\,4,\,6,\dots,\,2k}} b_i
        \end{equation*}
    \end{enumerate} 
    The length of each masked portion $b_i$ is randomly generated, while ensuring that the total masked proportion equals the hyperparameter $p$.
    \item \textbf{Signal warp}: This augmentation randomly stretch or squeeze portions of the EEG signal along the time axis, proceeding as follows:
    \begin{enumerate}[1.]
        \item The EEG signal is divided into multiple chunks (segments).
        \item Up to half of these segments are randomly chosen to be stretched, while the remaining segments are squeezed.
        \item Based on the random selection of the segments to squeeze or stretch, a non-uniform time grid is constructed. The values of this grid are defined according to the stretch and squeeze strengths hyperparameters.
        \item The EEG signal is interpolated onto this new grid using the Piecewise Cubic Hermite Interpolating Polynomial (PCHIP) \cite{pchip}.
        \item the new signal is treated as uniformly sampled between 0 and the sequence length L, and a final PCHIP interpolation is performed to resample the signal back onto the original time grid.
    \end{enumerate}
    Therefore, the augmented EEG sample $\tilde{\text{X}}$ is computed as:
    \begin{equation}
        \tilde{\text{X}} = \textit{pchip}( \space \textit{pchip}(\text{X}, \mathbf{t}_{old}, \mathbf{t}_{nu}), \mathbf{t}_{\text{u}}, \mathbf{t}_{old} \space)
    \end{equation}
    where:
    \begin{enumerate}[-]
        \item $\textit{pchip}(\text{X}, \mathbf{t}_{1}, \mathbf{t}_{2})$ is the function performing the interpolation of a signal from a time grid $\mathbf{t}_{1}$ to a time gird $\mathbf{t}_{2}$.
        \item $\mathbf{t}_{\text{old}}$ is the original time grid.
        \item $\mathbf{t}_{\text{nu}}$ is the non-uniform, warped time grid constructed at step 3.
        \item $\mathbf{t}_{\text{u}}$ is the uniformly sampled time grid used for the final interpolation at step 5.
    \end{enumerate}
    \item \textbf{Phase Randomizer}: This augmentation randomly perturbs the phase of the EEG signal. The augmented sample is computed by applying the transformation:
    \begin{equation}
            \tilde{\text{X}} = \text{Re}(\mathcal{F}^{-1}[\mathcal{F}[\text{X}]\odot (\mathbf{1}_c{e^{\circ js\boldsymbol{\phi}}}^T)])
    \end{equation}
    where:
    \begin{enumerate}[-]
        \item $\mathcal{F}[\text{X}]$ and $\mathcal{F}^{-1}[\text{X}]$ denote the discrete Fourier transform (DFT) and the inverse discrete Fourier transform (IDFT), respectively, applied to each channel of the EEG sample $\text{X}$.
        \item $e^{\circ X}\!=\!(e^{x_{ij}})$ is the Hadamard exponential, which computes the element-wise exponential of a matrix X.
        \item $\boldsymbol{\phi} \sim \mathcal{U}(0, 2\pi)^N$ is a vector of length $N$, where each element is independently drawn from a uniform distribution over $[0, 2\pi)$, used to randomize the phase of each EEG channel.
        \item $s$ is a scalar between 0 and 1 defining the strength of the phase perturbation.
    \end{enumerate}
    \item \textbf{Phase Swap}: Presented in \cite{phaseswap}, this data augmentation consists in merging the amplitude and phase components of biosignals from different sources to help the model learn their coupling. 
    Specifically, the amplitude and phase of two randomly selected EEG samples are extracted using the discrete Fourier transform. New samples are then generated by applying the inverse discrete Fourier transform, combining the amplitude from one sample with the phase from the other.
    This process is described by the following transformation:
    \begin{equation}
        \tilde{\text{X}}^{i}= \text{Re}\Big{(}\mathcal{F}^{-1} \Big{[} |\mathcal{F}[\text{X}^{i}]| \odot e^{\circ j\space\text{arg}(\mathcal{F}[\text{X}^{j}])} \Big{]} \Big{)}
    \end{equation}
    where:
    \begin{enumerate}[-]
        \item $|\mathcal{F}[\text{X}^{i}]|$ is the amplitude of the EEG sample $\text{X}^{i}$.
        \item $\text{arg}(\mathcal{F}[\text{X}^{j}])$ is the phase of the EEG sample $\text{X}^{j}$.  
    \end{enumerate}
\end{enumerate}
\WarningsOn

\noindent Whenever possible, the same data augmentation is applied to all channels of an EEG sample and broadcasted across the batch dimension to improve computational efficiency.
Additionally, data augmentation was performed during 75\% of the training iterations.
In other words, when a batch is created, there is a 75\% probability that it will be augmented with the selected data augmentations; otherwise, the identity function is applied.
This additional randomness helps balance the number of original and augmented samples provided to the model.
The optimal data augmentation for each model is reported in Section A.4 of the Supplementary Materials.

\subsubsection{Training Hyperparameters}
\label{subsubsec: train hyper}

A fixed random seed value of 42 was used to minimize randomness in the code and enhance reproducibility, as recommended in {\cite{DelPup2024a}}.
An analysis of how the random seed affects the accuracy estimation of N-LNSO cross-validation is provided in Section A.3 of the Supplementary Materials.
This analysis confirms that the study's conclusions are not influenced by the choice of the seed.

Model weights were initialized using Pytorch's default initialization settings and were trained with Adam optimizer ($\beta_{1} = 0.75$, $\beta_{2} = 0.999$, no weight decay) \cite{adam}.
The batch size was set to 64, and the initial learning rate was $2.5\cdot10^{-4}$.
An exponential scheduler with $\gamma = 0.99$ was used to reduce the learning rate after each epoch. The learning rate at the epoch $i$ is computed as:

\begin{equation}
    \text{lr}_i = \text{lr}_0 \times \gamma^i
\end{equation}

Binary cross entropy was used as loss function.
The maximum number of epochs was set to 300 to ensure convergence across all the selected models.
To prevent overfitting, early stopping with a patience of 20 epochs was implemented as an additional regularization method.
Early stopping can be integrated into N-LNSO cross-validation procedures, as each partition includes a unique triplet of training, validation, and test sets.

%% file: Ch3_Results.tex
\section{Results}
\label{sec: results}

This section presents the results of the comparative analysis of TransformEEG.
It is structured as follows:
\WarningsOff
\begin{enumerate}[\textbullet]
    \item Subsection \ref{subsec: baseline} presents the baseline results for TransformEEG and the seven models used for the comparison. These results are for a pipeline without data augmentation and threshold correction.
    \item Subsection \ref{subsec: DA effect} and subsection \ref{subsec: th effect} evaluate TransformEEG and the seven models used for the comparison within a richer training pipeline that includes data augmentation and threshold correction.
    \item Subsection \ref{subsec: model scalability} examines performance variations when the number of subjects is reduced, assessing the importance of data harmonization.
\end{enumerate}
\WarningsOn

During the presentation of the results, no statistical test outcomes will be reported.
As explained in \cite{NLOSO}, the Nested-Leave-N-Subjects-Out (N-LNSO) cross-validation procedure generates more reliable performance estimates.
However, it does not align with the key assumptions of statistical tests designed to identify significant differences in variance between models (e.g., Levene's test, Fligner-Killeen's test).
Specifically, sample independence is violated because different N-LNSO splits share overlapping training sets.
This violation increases the risk of Type I errors and can lead to misinterpretation of the results.

\subsection{Baseline comparison}
\label{subsec: baseline}

\begin{figure*}[!t]
    \centering
    \includegraphics[width=\textwidth]{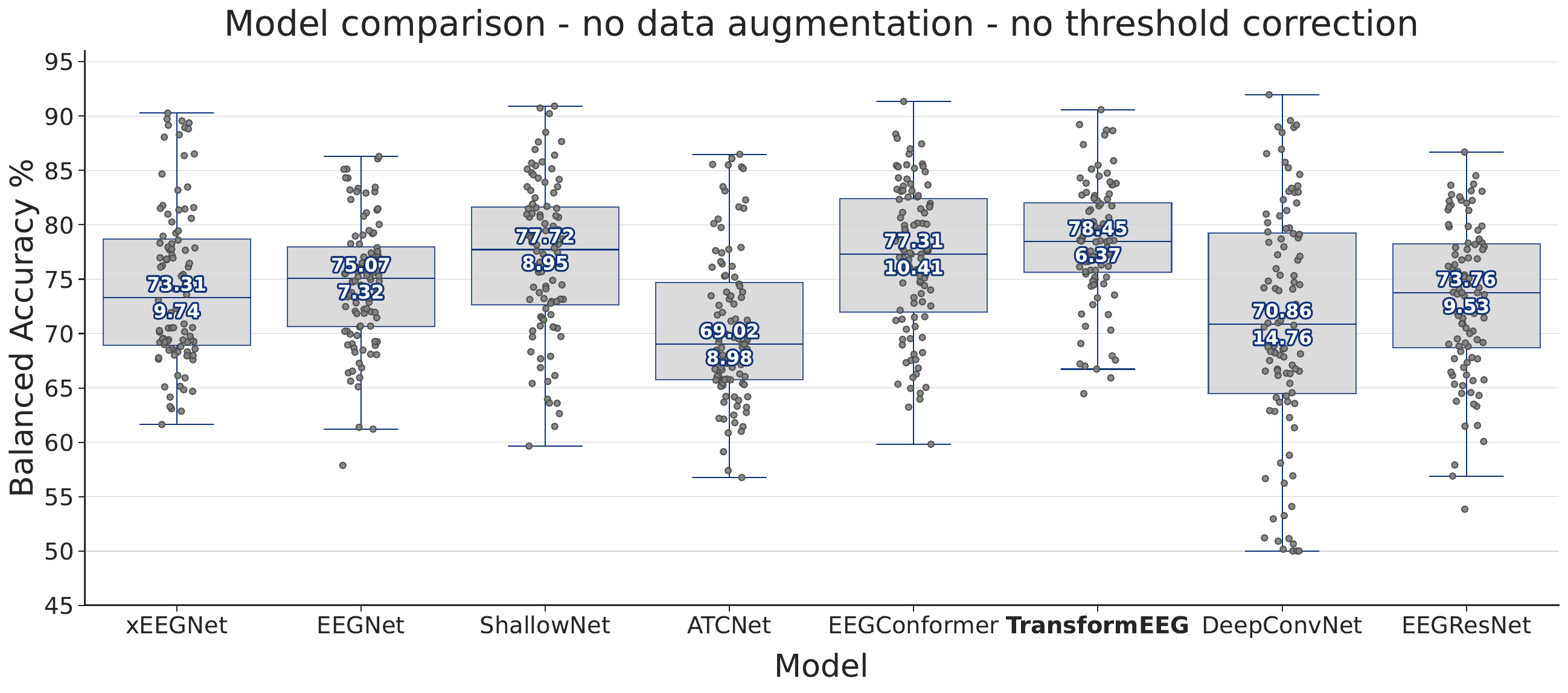}
    \caption{
    Performance comparison of the selected models using a 10-outer, 10-inner N-LNSO cross-validation scheme.
    Models are organized in ascending order based on the number of learnable parameters from left to right.
    All models were trained without data augmentation or threshold correction to serve as a baseline reference.
    TransformEEG ranks first in median balanced accuracy and interquartile range.
    It also achieved the highest minimum accuracy.
    These results demonstrate the superior generalizability of the proposed TransformEEG architecture for the investigated task.
    }
    \label{fig: baseline}
\end{figure*}

Figure \ref{fig: baseline} presents the results of TransformEEG compared to the other seven selected models.
No data augmentation or threshold correction was included at this stage to provide a baseline reference for subsequent analysis.
TransformEEG ranks first in terms of median balanced accuracy (78.45\%) and IQR (6.37\%).
It also achieved the highest minimum accuracy (64.46\%).
ShallowNet and EEGConformer achieve comparable median accuracies (77.72\% and 77.31\%, respectively), but they exhibit higher variability in the results.
Additionally, TransformEEG demonstrates a high median number of effective training epochs, defined as the number of epochs required to reach the best validation accuracy during training.
This metric indicates how well the model's reduction in training loss is accompanied by a corresponding reduction in validation loss, reflecting better generalization to unseen data.
A very low number of effective training epochs suggests that the model quickly reaches its optimal validation performance, which may be indicative of overfitting tendencies and lower generalization capabilities. On the contrary, a higher number of epochs can imply better generalization, as the model continues to improve over more training iterations.
With a median of 21 effective training epochs, TransformEEG ranks among the best models, second only to the minimalist xEEGNet, which has a median of 37 epochs.
This result demonstrates that, despite having \num{210561} learnable parameters, TransformEEG does not rapidly overfit the training set as some comparable transformer architectures do, such as EEGConformer, which has a median of only two effective training epochs.

\subsection{Comparison including data augmentation}
\label{subsec: DA effect}

\begin{figure*}[!t]
    \centering
    \includegraphics[width=\textwidth]{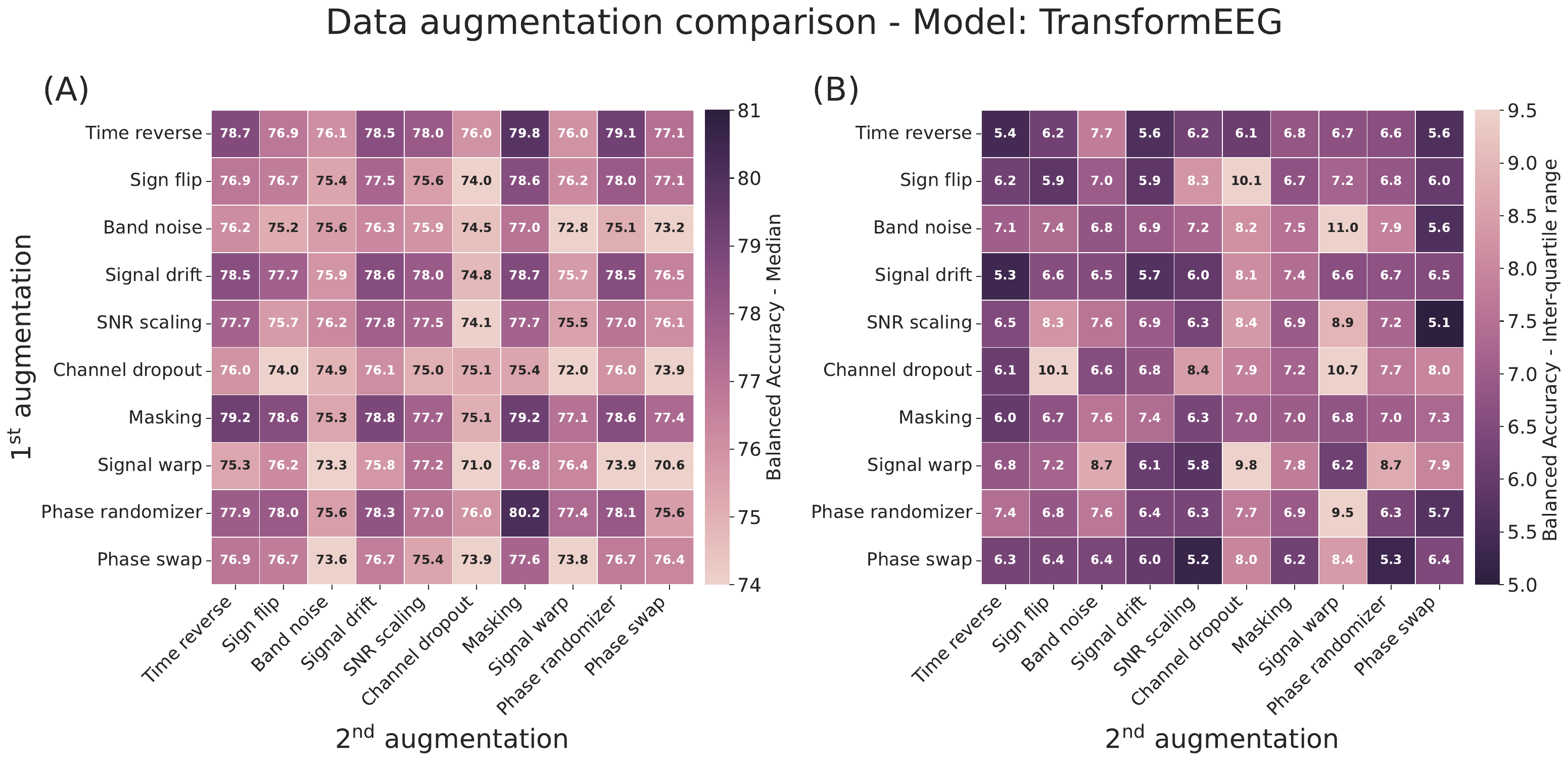}
    
    \includegraphics[width=0.9\textwidth]{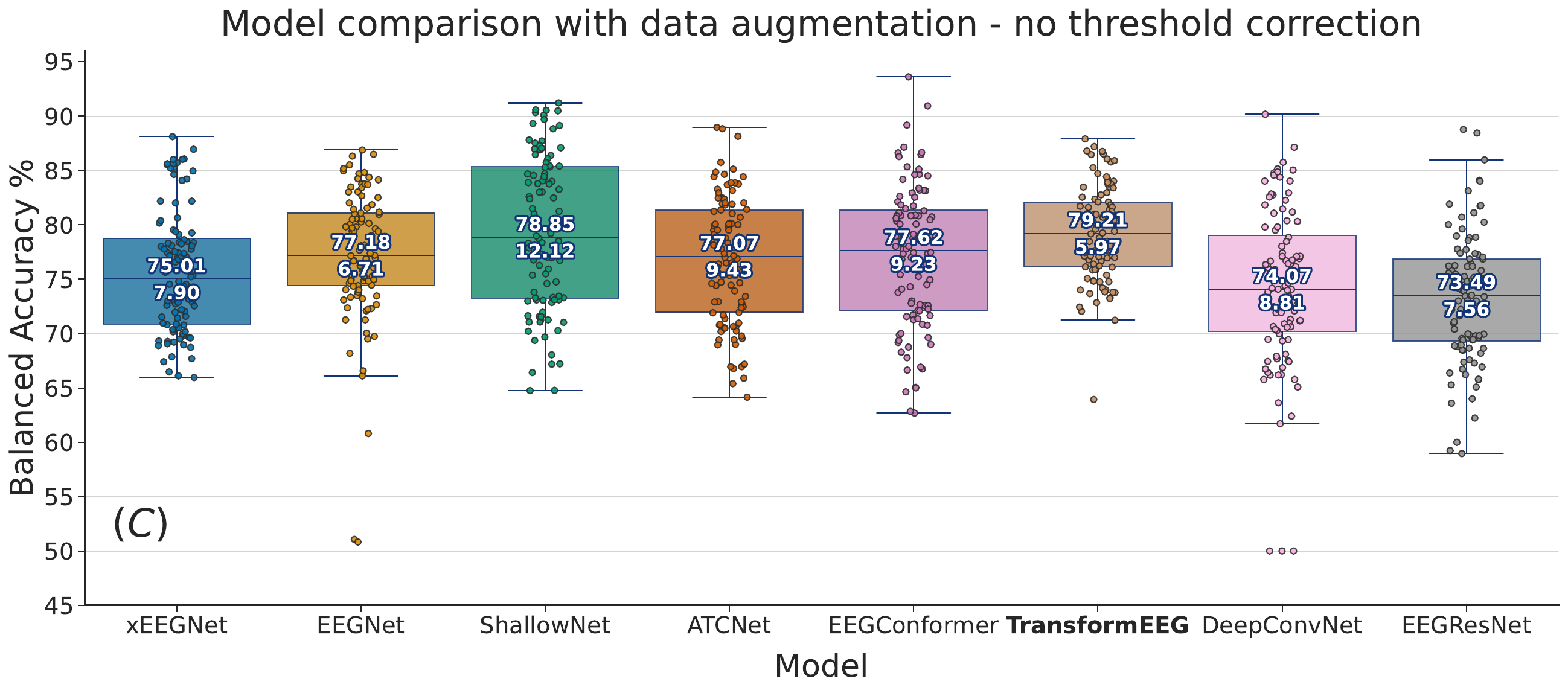}
    \caption{
    Effect of data augmentation on the performance of TransformEEG and the other selected EEG-based deep learning models.
    Panels A and B display the median and interquartile range of balanced accuracy, respectively, when different data augmentation compositions are incorporated during the training of TransformEEG.
    For each augmentation combination, an N-LNSO cross-validation was performed.
    Panel C shows the results of a 10-outer 10-inner N-LNSO performance comparison between the selected models when the optimal data augmentation composition for each model is applied during training.
    Models are organized in ascending order based on the number of learnable parameters from left to right.
    TransformEEG ranks first in median balanced accuracy and interquartile range.
    These results improves baseline values and underscores the potential benefit of integrating properly tuned data during training.
    }
    \label{fig: da_comparison}
\end{figure*}

This subsection examines how performance varies when an optimally selected data augmentation is incorporated into the training pipeline.
It highlights how TransformEEG maintains leading performance within a fair comparison and, more generally, how this step can enhance performance and reduce variability across the models considered.

Figure \ref{fig: da_comparison}, Panels A and B, show the results of TransformEEG when different data augmentation compositions are applied during training, following the procedure described in subsection \ref{subsubsec: data augmentation}.
A summary of the results from the same analysis performed on the other seven models is presented in Section A.4 of the Supplementary Materials.
Focusing on TransformEEG, several compositions identify as a good candidate.
In particular, using ``time reverse'' in combination with either ``masking'' or ``phase swap'' ensures a good balance between the median and IQR of the balanced accuracy.
The combination of time reverse with phase swap results in a decreased IQR (from 6.4\% to 5.6\%) at the cost of a slight decrease of the median value (-1.3\%).
However, this composition does not alter the power spectral density (PSD) of the input EEG window, which can be advantageous in scenarios where spectral information are combined with temporal patterns to enrich the quality of the representations (see Section \ref{sec: discussion}).
Masking with time reverse allows for a decrease of the IQR (from 6.4\% to 6.0\%) and a slight increase in the median value (+0.6\%), achieving the highest ARIS. 
Therefore, it was selected and tested against the optimal combination of other models.

Figure \ref{fig: da_comparison}, Panel C, shows the results of TransformEEG compared to the seven other selected deep learning models when optimal data augmentation is applied during training.
TransformEEG ranks first in terms of median balanced accuracy (79.21\%) and interquartile range (5.97\%).
As in the previous subsection, ShallowNet and EEGConformer achieve comparable median accuracies (78.85\% and 77.62\%, respectively), but they exhibit higher variability in their results.
These results improve upon the baseline values reported in subsection \ref{subsec: baseline} and confirm that data augmentation can be effectively integrated into EEG deep learning pipelines to enhance the model's generalizability.
Additionally, if the single outlier is discarded from the TransformEEG's N-LNSO set of training instances, the highest minimum balanced accuracy would raise from 64.5\% to 71.2\%.
In fact, TransformEEG is the only model to have nearly all test accuracies (99 out of 100) above 70\%, achieving the lowest $[1^{st}\!-\!99^{th}]$ percentile range (16.05\%).
Low variability revealed that TransformEEG can achieve competitive performance with more consistent results.

Considering the entire set of results, all the models achieved an increase in the median balanced accuracy.
The interquartile range also decreases in most of the models, with ShallowNet being the only exception (from the baseline 8.95\% to 12.12\%).
Despite the increment in the IQR, ShallowNet's $[1^{st}\!-\!99^{th}]$ percentile range improved from 29.90\% to 25.81\%.
These improvements over baseline values underscore the potential of integrating properly tuned data augmentations during the training of EEG deep learning models.

\subsection{Comparison including threshold correction}
\label{subsec: th effect}

Table \ref{tab: summary results} shows the results of TransformEEG against the other selected models when the threshold correction method is incorporated to the training pipeline.
This approach consists in adjusting the default classification threshold of 0.5 by searching for an optimal value that maximize the balanced accuracy of the validation set.
This adjustment serves as a minor correction in case the trained model generates too aggressive or too conservative predictions, particularly when the class ratio is slightly unbalanced.
Using the validation data avoids introducing data leakage, as test data cannot be used to inflate the results.

Even in this scenario, TransformEEG ranks first in terms of median and IQR of the balanced accuracy. 
It is the sole model to achieve a median balanced accuracy higher than 80\% while also reducing the IQR from the baseline value of 6.37\% to 5.74\%.
ShallowNet maintains a comparable median accuracy (79.97\%) but with a higher variability of results.
This denotes how EEG-specific deep learning models trained in combination with carefully selected data augmentations and label assignment methods can effectively address generalizability issues.

\subsection{Model scalability}
\label{subsec: model scalability}

This subsection examines performance variations when the amount of training data is reduced.
It highlights generalizability limitations common to all architectures, including smaller models, which are favored in this type of analysis due to their lower number of parameters.

Previous subsections (\ref{subsec: baseline}, \ref{subsec: DA effect}, and \ref{subsec: th effect}) demonstrate the efficacy of TransformEEG, which outperforms other models in terms of the interquartile range and the $[1^{st}\!-\!99^{th}]$ percentile range of the balanced accuracy. 
These findings are consistent across different metrics, as confirmed in Section A.2 of the Supplementary Materials.
However, training deep learning models such as TransformEEG often heavily depends on the amount of training data available.
Table \ref{tab: summary results} illustrates how performance varies when the number of subjects decreases from 290 to 81, using data from the ds002778 and ds003490 datasets, as in \cite{DelPup2024b}.
Focusing on the baseline results (Table \ref{tab: summary results}, ``B'' rows), performance variability across all investigated models increases substantially, even in lighter models such as xEEGNet (from 9.74\% to 18.02\%).
TransformEEG's interquartile range raises similarly, despite remaining one of the lowest among all the investigated models.
Its median balanced accuracy also drops by 6.01\%, ranking below other lighter models such as ShallowNet.
When training with only two datasets, adding data augmentation or threshold correction does not produce the same benefits observed when using data from all four datasets.
This result emphasizes the importance of standardizing and aggregating EEG recordings from multiple centers to facilitate more effective learning of disease-specific EEG features and to ensure consistent results across different splits.

\begin{table*}
\centering
\caption{
Balanced accuracy of the selected models trained with different training modalities and varying number of datasets.
Acronyms in the ``Training Method" column are: B (baseline), DA (data augmentation), T (threshold correction). IQR indicates interquartile range.
}
\begin{tabular}{cclcccccc}
\toprule
\multicolumn{1}{c}{\multirow{3}{*}{Models}} & \multicolumn{1}{c}{\multirow{3}{*}{\makecell{\#\\Param}}} & \multicolumn{1}{c}{\multirow{3}{*}{\makecell{Training\\Method}}} & \multicolumn{6}{c}{Balanced Accuracy} \\ \cline{4-9}
 & & & \multicolumn{3}{c}{2 Datasets} & \multicolumn{3}{c}{4 Datasets} \\
\cline{4-9}
 &  &  & Median & IQR & $[1^{st}\!-\!99^{th}]$ & Median & IQR & $[1^{st}\!-\!99^{th}]$ \\ \midrule
\multirow{3}{*}{xEEGNet \cite{xeegnet}} & \multirow{3}{*}{\num{245}} & B & 67.99 & 18.02 & 47.00 & 73.31 & 9.74 & 26.90 \\
 &  & B+DA & \textbf{77.28} & 24.83 & 51.01 & 75.01 & 7.90 & 20.86 \\
 &  & B+DA+T & \textbf{77.35} & 22.69 & 48.88 & 78.03 & 9.84 & 20.39 \\
 \midrule
\multirow{3}{*}{EEGNet \cite{eegnet}} & \multirow{3}{*}{\num{2609}} & B & 70.03 & 20.96 & 44.76 & 75.07 & 7.32 & 24.90 \\
 &  & B+DA & 70.58 & 18.01 & 48.98 & 77.18 & 6.71 & 35.44 \\
 &  & B+DA+T & 67.42 & 20.20 & 48.73 & 78.44 & 6.88 & 25.10 \\
 \midrule
\multirow{3}{*}{ShallowNet \cite{shallow}} & \multirow{3}{*}{\num{57441}} & B & \textbf{78.73} & 18.18 & \textbf{41.73} & 77.72 & 8.95 & 29.90 \\
 &  & B+DA & 71.49 & 18.48 & 48.06 & 78.85 & 12.12 & 25.81 \\
 &  & B+DA+T & 72.29 & 16.97 & 49.07 & 79.91 & 11.00 & 25.44 \\
 \midrule
\multirow{3}{*}{ATCNet \cite{atcnet}} & \multirow{3}{*}{\num{146629}} & B & 72.73 & 15.97 & 44.83 & 69.02 & 8.98 & 28.71\\
 &  & B+DA & 73.02 & 17.56 & 46.95 & 77.07 & 9.43 & 23.46 \\
 &  & B+DA+T & 71.89 & 17.32 & 49.14 & 76.85 & 9.66 & 23.71 \\
 \midrule
\multirow{3}{*}{EEGConformer \cite{eegconformer}} & \multirow{3}{*}{\num{191153}} & B & 75.98 & 15.83 & 45.14 & 77.31 & 10.41 & 25.18 \\
 &  & B+DA & 75.46 & 16.67 & 49.56 & 77.62 & 9.23 & 28.13 \\
 &  & B+DA+T & 75.77 & \textbf{14.74} & 50.00 & 77.72 & 9.59 & 27.71 \\ 
 \midrule
\multirow{3}{*}{TransformEEG} & \multirow{3}{*}{\num{210561}} & B & 72.09 & 15.93 & 44.83 & \textbf{78.45} & \textbf{6.37} & \textbf{23.33} \\
 &  & B+DA & 69.92 & \textbf{14.39} & 44.62 & \textbf{79.21} & \textbf{5.97} & \textbf{16.05} \\
 &  & B+DA+T & 70.62 & 15.07 & \textbf{43.42} & \textbf{80.10} & \textbf{5.74}  & \textbf{18.21} \\
 \midrule
\multirow{3}{*}{DeepConvNet \cite{shallow}} & \multirow{3}{*}{\num{287901}} & B* & 50.09 & 0.64 & 37.17 & 70.86 & 14.76 & 39.60 \\
 &  & B+DA & 67.97 & 29.47 & \textbf{40.09} & 74.07 & 8.81 & 37.16 \\
 &  & B+DA+T & 74.14 & 32.49 & 46.62 & 75.64 & 9.35 & 36.04 \\
 \midrule
\multirow{3}{*}{EEGResNet \cite{resnet2}} & \multirow{3}{*}{\num{1337665}} & B & 73.12 & \textbf{13.65} & 47.71 & 73.76 & 9.53 & 27.68 \\
 &  & B+DA & 74.24 & 16.86 & 44.34 & 73.49 & 7.56 & 29.21 \\
 &  & B+DA+T & 74.96 & 17.80 & 48.29 & 75.10 & 6.92 & 25.10 \\
 \bottomrule
 \multicolumn{9}{l}{*Baseline results of DeepConvNet trained with two datasets were ignored because the model collapsed.}
\end{tabular}
\label{tab: summary results}
\end{table*}

%% file: Ch4_Discussions.tex
\section{Discussion}
\label{sec: discussion}

EEG-based deep learning models often exhibit high variability, which can only be assessed through the use of appropriate model evaluation techniques \cite{NLOSO}.
This aspect is particularly relevant in pathology detection tasks, where the direct association between patient ID and health status (class label) can strongly influence training dynamic, serving as a shortcut for minimizing the training loss \cite{LOSO2}.
Nested approaches, such as the N-LNSO, offer more reliable performance estimates but also reveal strong variability related to changes in data splits.
This variability complicates the comparison of different EEG deep learning models, as median accuracy values become less informative due to the wide range of accuracies observed across different splits.

TransformEEG was specifically designed to advance current generalizability challenges in EEG pathology detection, focusing on Parkinson's disease.
It features a depthwise convolutional tokenizer that specializes in generating tokens describing local time segments of the EEG window with channel-specific features.
This tokenizer differs from EEGNet-like convolutional blocks and enables more effective feature mixing operations within the transformer's self-attention layers, as presented in subsection \ref{subsec: transformeeg}.

Baseline results in subsection \ref{subsec: baseline} show that TransformEEG ranks first in terms of the median (78.45\%) and IQR (6.37\%) balanced accuracy.
It also achieved the highest minimum accuracy (64.5\%).
These results improve further if additional regularization techniques, such as data augmentation, are applied during training.
After applying an optimal data augmentation composition for each model and adjusting the classification threshold, TransformEEG achieves a median balanced accuracy of 80.10\% and an IQR of 5.74\%, remaining the best among the models investigated (see subsection \ref{subsec: th effect}).

The strength of TransformEEG lies in the clear and complementary roles assigned to each of its modules.
For example, the depthwise convolutional tokenizer does not perform feature mixing between EEG channels, unlike other EEG transformer models \cite{eegconformer}.
Instead, it specializes in extracting representations based on local temporal patterns within each individual EEG channel.
This design allows the transformer encoder, where most of the model's parameters reside, to be the sole module responsible for feature mixing across channels.
The combination of convolutional and attention layers enhances the network's ability to capture both local and global patterns within the EEG window, leading to more consistent performance across different N-LNSO splits.
This consistency highlights the better capability of the model to generalize to diverse data and partitions.

To enable attention layers to capture long-range temporal patterns without overfitting the training set, it is essential to use a sufficient number of EEG recordings from different subjects.
Results in subsection \ref{subsec: model scalability} investigate how model performance changes when the number of subjects is reduced from 290 to 81, replicating the experimental setting presented in \cite{DelPup2024b}.
When only 81 subjects are used, performance variability across all investigated models increases drastically.
TransformEEG was not excluded by this trend.
The interquartile range increase of 9.56\% and the median balanced accuracy drop of 6.01\%.
This result suggests that a larger number of subjects is necessary to enable the effective training of EEG deep learning architectures.
It also emphasizes the need for creating novel large-scale, multi-center datasets. 

As discussed in recent works, EEG deep learning classifiers can easily exploit the unique association between labels and participant IDs as a shortcut to minimize the training loss \cite{sample_leakage_2, NLOSO}.
This association induces severe overfitting and prevents the model from learning disease-specific representations.
Inter-subject heterogeneity is therefore a primary source of variability, and the most effective way to address this challenge is by aggregating data from different sources, as done in this study.
Harmonization tools like BIDSAlign \cite{BIDSAlign} can preprocess and standardize EEG recordings from multiple centers, enabling the training of more robust EEG deep learning models and facilitating fair benchmarking.
Open platforms like OpenNeuro provide access to a large number of EEG datasets.
Despite potential differences in experimental paradigms (e.g., resting-state vs. task-based) or population characteristics, these datasets can improve model generalizability, particularly if label-free paradigms such as self-supervised learning are used \cite{rafieissl}.
This approach allows for the exploration and comparison of alternative architectures such as transformers, known to require large amounts of data.
Ultimately, these strategies can advance the field and enhance the reliability of EEG-based deep learning decision support systems.

TransformEEG demonstrates promising results in Parkinson's disease detection and lays the foundation for addressing generalizability issues in EEG-based deep learning pathology classification tasks \cite{sample_leakage_2}.
However, model generalizability depends not only on the architecture design, but also on how data are preprocessed, partitioned, and used to train and evaluate the model.
%
TransformEEG is a model based on temporal analysis.
It takes an EEG window in the time domain as input and outputs the probability that the window belongs to one of the investigated diseases.
While this approach aligns with an optimal engineering solution for addressing the challenges of small dataset sizes and GPU memory limitations, it does not necessarily represent the best approach for disease detection.
In pathology classification tasks, it is of greater clinical interest to determine whether the entire EEG originates from an individual with the target disease, rather than focusing solely on a single time window.
Therefore, voting or aggregation procedures should be considered and carefully integrated into the evaluation process.
Not all EEG segments are equally informative of a neurological disease, especially when short window lengths are used.
Artifacts or involuntary movements can obscure disease-specific brain dynamics and alter the segment's power spectral density.
Aggregation procedures can help address this issue by smoothing disease detection predictions across all segments from the same EEG recording. 
See subsection A.7.1 of the supplementary materials for a preliminary analysis of aggregation procedures.

Contrary to other models such as xEEGNet, TransformEEG does not process the EEG window in the time domain to specifically extract features that are linearly proportional to the mean power of EEG bands across channels.
While this design choice enhances the performance of the model, it would be of great clinical interest to assess whether combining both temporal and spectral information could further improve model generalizability.
Researchers have already explored approaches that integrate temporal and spectral information to boost model performance, typically by extracting the spectrogram of an EEG window \cite{dicenet}.
This is also supported in \cite{babiloni}, where it is stated that the integration of spectral information, such as the power spectral density (PSD), can aid in the assessments of certain neurodegenerative disorders.
If resting-state data are used, as in this study, a potential strategy can involve incorporating the PSD of the EEG segment in a separate branch to generate an additional set of features, or tokens if the model includes transformer blocks.
Temporal and spectral information can then be combined with mid- or late-fusion strategies.
See subsection A.7.2 of the supplementary materials for a preliminary analysis on the integration of the PSD within the TransformEEG model.

Results in Section \ref{sec: results} demonstrate that TransformEEG can learn features that improve accuracy on unseen subjects.
This suggests that the learned representations might capture more effectively disease-specific characteristics rather than biometric properties of the EEG signal.
However, this study focuses on Parkinson's disease detection using a binary classification task.
Parkinson's disease is a complex and heterogeneous disorder characterized by a variety of symptoms that can lead to different disease subtypes, often influenced by the severity of cognitive decline. 
Differentiating these subtypes is of great clinical importance, as it is essential for developing personalized therapies.
The availability of new, large, dedicated datasets will facilitate their integration with existing data and support the investigation of deep learning approaches to automatically identify PD subcategories within a multi-class classification framework.
This aggregation can address current challenges, such as overlapping clinical symptoms and involved brain regions \cite{multiclass}, and enhance the baseline performances reported in recent studies on similar classification problems \cite{DelPup2024b, xeegnet}.
Such advancements, along with the future directions discussed earlier, will enhance the reliability of EEG-based deep learning models for pathology detection.
They will also facilitate the integration of these systems into real-world clinical scenarios, ultimately supporting clinicians and patients more effectively.

%% file: Ch5_Conclusions.tex
\section{Conclusion}
\label{sec: conclusion}

This work introduces TransformEEG, a novel hybrid convolutional-transformer model designed to address generalizability issues in Parkinson's disease detection from EEG data.
TransformEEG features a depthwise convolutional tokenizer that specializes in generating tokens representing temporally local segments of the EEG window with channel-specific features.
This module enables more effective feature mixing within the transformer encoder, which is included to capture long-range temporal patterns.
The model was evaluated using harmonized EEG recordings from four publicly available datasets, comprising 290 subjects (140 Parkinson's Disease individuals and 150 healthy controls), through a Nested-Leave-N-Subjects-Out cross-validation scheme.
When trained with a sufficient amount of data, TransformEEG achieved promising results, ranking first among seven other consolidated EEG deep learning models in terms of median balanced accuracy (80.10\%), interquartile range (5.74\%).
These findings demonstrate that designing deep learning models tailored to EEG data and evaluating them on large, multi-center datasets can help address generalizability challenges, despite the path to solutions that generalize to clinical settings remains an open challenge for the global scientific community.

%% file: supplementary.tex
\appendix

\section{Supplementary Materials}

The following subsections provides supplementary materials for the research study ``TransformEEG: Towards Improving Model Generalizability in Deep Learning-based EEG Parkinson's Disease Detection".

\subsection{Nested-Leave-N-Subjects-Out scheme}

\begin{figure}[!b]
    \centering
    \includegraphics[width=\linewidth]{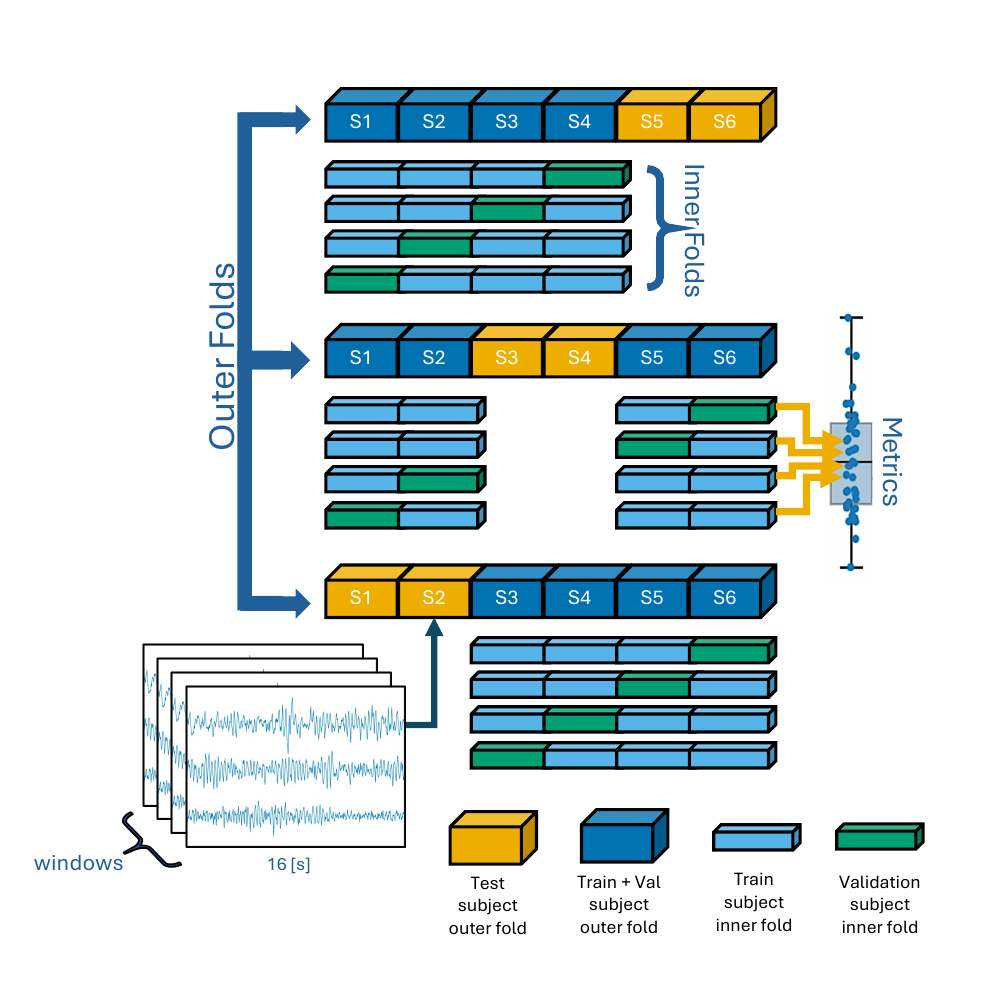}
    \caption{
    Scheme of the N-LNSO cross-validation method proposed in Del Pup \etal, 2025. N-LNSO adds an additional nested level to Leave-N-Subjects-Out approaches to generate a set of train-validation-test splits.
    Models are trained on the training set, monitored with early stopping techniques on the validation set, and evaluated on test set.
    This study uses ten outer and ten inner folds for a total of 100 splits.
    The generated ensemble of performances is used to compare deep learning models.
    }
    \label{fig: nlnso scheme}
\end{figure}

This section provides additional details about the Nested-Leave-N-Subjects-Out (N-LNSO) cross-validation scheme used in this study. 
Figure \ref{fig: nlnso scheme} schematizes how it works.
As explained in Del Pup \etal, 2025, N-LNSO can be considered as a concatenation of two different subject-based cross-validations to create a set of multiple train-validation-test splits.
The first cross-validation determines the subjects to use as test set, which are called outer folds in the figure.
The second further partitions the remaining subjects to create the training and validation sets (inner folds).
Models are trained with the training set, monitored with the validation set using early stopping, and evaluated with the test set.
N-LNSO generates triplets of independent test sets composed of different subjects, ensuring a correct and fair comparison between models without any form of data leakage.
N-LNSO generates an ensemble of of $N_{outer} \times N_{inner}$ accuracies, which are used to compare the models and their consistency across different partitions.

\subsection{Summary results with other metrics}
\label{supplement: metric}

This section summarizes the results of the N-LNSO cross-validations runs using additional performance metrics.
In particular, Table {\ref{tab: f1_and_kappa}} reports the median, interquartile range, and $[1^{st} - 99^{th}]$ percentile range for the weighted F1-score and Cohen's Kappa.
TransformEEG achieves the best IQR and percentile range for both metrics.
Notably, ShallowNet reaches comparable median values and has the highest Cohen's Kappa (0.587 in baseline + data augmentation), while TransformEEG has the highest F1-score (0.811 in baseline + data augmentation).
These results demonstrate that TransformEEG can achieve competitive performance while providing more consistent results across splits.

\begin{table*}[!b]
\centering
\caption{
Weighted F1-score and Cohen's Kappa for the selected models trained on four datasets with different training modalities.
Acronyms in the ``Training Method" column are: B (baseline), DA (data augmentation), T (threshold correction). IQR indicates interquartile range.
}
\begin{tabular}{cclcccccc}
\toprule
\multicolumn{1}{c}{\multirow{3}{*}{Models}} & \multicolumn{1}{c}{\multirow{3}{*}{\makecell{\#\\Param}}} & \multicolumn{1}{c}{\multirow{3}{*}{\makecell{Training\\Method}}} & \multicolumn{6}{c}{10-Outer 10-Inner N-LNSO - 4 Datasets} \\ \cline{4-9}
 & & & \multicolumn{3}{c}{F1-score weighted} & \multicolumn{3}{c}{Cohen's Kappa} \\
\cline{4-9}
 &  &  & Median & IQR & $[1^{st}\!-\!99^{th}]$ & Median & IQR & $[1^{st}\!-\!99^{th}]$ \\ \midrule
\multirow{3}{*}{xEEGNet \cite{xeegnet}} & \multirow{3}{*}{\num{245}} & B & 0.782 & 0.092 & 0.247 & 0.469 & 0.197 & 0.492 \\
 &  & B+DA & 0.790 & 0.070 & 0.183 & 0.513 & 0.145 & 0.379 \\
 &  & B+DA+T & 0.784 & 0.065 & 0.191 & 0.541 & 0.147 & 0.406 \\
 \midrule
\multirow{3}{*}{EEGNet \cite{eegnet}} & \multirow{3}{*}{\num{2609}} & B & 0.775  & 0.075 & 0.223 & 0.495 & 0.120 & 0.452 \\
 &  & B+DA &  0.794 & 0.061 & 0.267 & 0.537 & 0.103 & 0.661 \\
 &  & B+DA+T & 0.781 & 0.052 & 0.270 & 0.534 & 0.097 & 0.536 \\
 \midrule
\multirow{3}{*}{ShallowNet \cite{shallow}} & \multirow{3}{*}{\num{57441}} & B & 0.797 & 0.072 & 0.247 & 0.555 & 0.155 & 0.503 \\
 &  & B+DA & 0.810 & 0.083 & 0.267  & \textbf{0.587} & 0.186 & 0.444 \\
 &  & B+DA+T & \textbf{0.810} & 0.079 & 0.218 & 0.580 & 0.155 & 0.454 \\
 \midrule
\multirow{3}{*}{ATCNet \cite{atcnet}} & \multirow{3}{*}{\num{146629}} & B & 0.742 & 0.074 & 0.276 & 0.407 & 0.152 & 0.530 \\ 
 &  & B+DA & 0.786 & 0.056 & 0.223 & 0.530 & 0.150 & 0.463 \\
 &  & B+DA+T & 0.784 & 0.060 & 0.278 & 0.532 & 0.156 & 0.503 \\
 \midrule
\multirow{3}{*}{EEGConformer \cite{eegconformer}} & \multirow{3}{*}{\num{191153}} & B & 0.787 & 0.073 & 0.219 & 0.540 & 0.150 & 0.488 \\
 &  & B+DA & 0.783 & 0.066 & 0.244 & 0.534 & 0.148 & 0.475 \\
 &  & B+DA+T & 0.788 & 0.068 & 0.235 & 0.543 & 0.144 & 0.498 \\
 \midrule
\multirow{3}{*}{TransformEEG} & \multirow{3}{*}{\num{210561}} & B & \textbf{0.800} & \textbf{0.050} & \textbf{0.184} & \textbf{0.557} & \textbf{0.088} & \textbf{0.374} \\
 &  & B+DA & \textbf{0.811} & \textbf{0.041} & \textbf{0.144} & 0.574 & \textbf{0.097} & \textbf{0.329} \\
 &  & B+DA+T & 0.806 & \textbf{0.040} & \textbf{0.142} & \textbf{0.586} & \textbf{0.092}  & \textbf{0.283} \\
 \midrule
\multirow{3}{*}{DeepConvNet \cite{shallow}} & \multirow{3}{*}{\num{287901}} & B & 0.760 & 0.129 & 0.453 & 0.440 & 0.272 & 0.776 \\
 &  & B+DA & 0.771 & 0.064 & 0.327 & 0.494 & 0.127 & 0.743 \\
 &  & B+DA+T & 0.763 & 0.078 & 0.325 & 0.502 & 0.158 & 0.649 \\
 \midrule
\multirow{3}{*}{EEGResNet \cite{resnet2}} & \multirow{3}{*}{\num{1337665}} & B & 0.771 & 0.087 & 0.266 & 0.483 & 0.191 & 0.517 \\
 &  & B+DA & 0.773 & 0.081 & 0.372 & 0.497 & 0.144 & 0.569 \\
 &  & B+DA+T & 0.770 & 0.071 & 0.292 & 0.493 &  0.135 & 0.516\\
 \bottomrule
\end{tabular}
\label{tab: f1_and_kappa}
\end{table*}

\subsection{Changing the random seed}
\label{supplement: seed}

As stated in subsection 2.4.3, a fixed random seed value of 42 was used to minimize randomness in the code and improve reproducibility.
To evaluate whether TranformEEG achieves sufficiently consistent results, multiple N-LNSO cross-validation runs using different seed values were performed.
Table {\ref{tab: seed_effect}} reports median, interquartile range, and $[1^{st} - 99^{th}]$ percentile range for the balanced accuracy with the seeds 1, 12, and 42.
The results show that, although there is some variability in N-LNSO performance due to randomness, TransformEEG consistently performs better than the other models in terms of variability.

\begin{table}[!b]
\centering
\caption{
Balanced accuracy of models trained on a baseline pipeline without data augmentation or threshold correction, evaluated across different seed numbers.
}
\begin{tabular}{ccccc}
\toprule
\multicolumn{1}{c}{\multirow{2}{*}{Models}} & \multicolumn{1}{c}{\multirow{2}{*}{seed}} & \multicolumn{3}{c}{Balanced Accuracy} \\ \cline{3-5}
 & & Median & IQR & $[1^{st}\!-\!99^{th}]$ \\ \midrule
\multirow{3}{*}{xEEGNet} 
 & 1 & 64.85 & 12.21 & 30.96 \\
 & 12 & 70.41 & 13.37 & 32.87 \\
 & 42 & 73.31 & 9.74 & 26.90 \\
 \midrule
\multirow{3}{*}{EEGNet} 
 & 1 & 71.55 & 10.88 & 32.29 \\
 & 12 & 73.67 & 8.47 & 25.43 \\
 & 42 & 75.07 & 7.32 & 24.90 \\\midrule
\multirow{3}{*}{ShallowNet} 
 & 1 & 77.49 & 9.33 & 30.89 \\
 & 12 & 77.19 & 10.01 & 33.32 \\
 & 42 & 77.72 & 8.95 & 29.90 \\
 \midrule
\multirow{3}{*}{ATCNet}
 & 1 & 71.92 & 7.32 & 26.07 \\
 & 12 & 71.48 & 9.58 & 24.16 \\
 & 42 & 69.02 & 8.98 & 28.71 \\
 \midrule
\multirow{3}{*}{EEGConformer}
 & 1 & 76.79 & 9.19 & 30.72 \\
 & 12 & 77.21 & 10.12 & 31.55 \\
 & 42 & 77.31 & 10.41 & 25.18 \\
 \midrule
\multirow{3}{*}{TransformEEG}
 & 1 & 77.33 & 6.28 & 23.86 \\
 & 12 & 76.10 & 8.18 & 22.32 \\
 & 42 & 78.45 & 6.37 & 23.33 \\
 \midrule
\multirow{3}{*}{DeepConvNet}
 & 1 & 70.76 & 12.38 & 40.98 \\
 & 12 & 67.69 & 15.13 & 37.88 \\
 & 42 & 70.86 & 14.76 & 39.60 \\
 \midrule
\multirow{3}{*}{EEGResNet} 
 & 1 & 74.34 & 8.22 & 28.95 \\
 & 12 & 74.07 & 8.47 & 26.28 \\
 & 42 & 73.76 & 9.53 & 27.68 \\
 \bottomrule
\end{tabular}
\label{tab: seed_effect}
\end{table}

\subsection{Choice of the data augmentation}
\label{subsec: da choice}

\begin{table*}[!H]
\caption{Top three data augmentations for each model. The median and interquartile range (IQR) of balanced accuracy are based on a 10-Outer, 5-Inner N-LNSO cross-validation.}
\label{tab: da selection}
\begin{tabular}{clccc}
\toprule
\multirow{2}{*}{Models} & \multicolumn{1}{c}{\multirow{2}{*}{\makecell[c]{Top 3\\data augmentations.}}} & \multicolumn{2}{c}{Balanced Accuracy} \\ \cline{3-5} 
 &  & Median & IQR & ARIS \\ \midrule
\multirow{4}{*}{xEEGNet} & Baseline & 73.51 & 10.98 & - \\
 & signal drift + phase swap & 76.08 & 6.86 & $1.31\cdot10^{-2}$ \\
 & masking + phase swap & 76.31 & 7.29 & $1.28\cdot10^{-2}$ \\
 & phase Swap + sign flip & 75.09 & 5.66 & $1.04\cdot10^{-2}$ \\ \midrule
\multirow{4}{*}{EEGNet} & Baseline & 74.41 & 6.61 & - \\
 & channel dropout + band noise & 75.69 & 6.05 & $1.46\cdot10^{-3}$ \\
 & band noise & 74.99 & 5.85 & $9.00\cdot10^{-4}$ \\
 & band noise + time reverse & 74.73 & 5.88 & $4.70\cdot10^{-4}$ \\ \midrule
\multirow{4}{*}{ShallowNet} & Baseline & 77.03 & 11.37 & - \\
 & channel dropout + masking & 80.40 & 8.02 & $1.29\cdot10^{-2}$\\
 & channel dropout + SNR scaling & 79.65 & 7.76 & $1.08\cdot10^{-2}$ \\
 & channel dropout & 78.48 & 6.99 & $7.26\cdot10^{-3}$ \\ \midrule
\multirow{4}{*}{ATCNet} & Baseline & 69.40 & 10.76 & - \\
 & phase swap + masking & 75.18 & 5.85 & $3.81\cdot10^{-2}$ \\
 & time reverse + phase swap & 75.04 & 6.40 & $3.30\cdot10^{-2}$ \\
 & SNR scaling + phase swap & 74.64 & 6.14 & $3.24\cdot10^{-2}$ \\ \midrule
\multirow{4}{*}{EEGConformer} & Baseline & 77.39 & 7.58 & - \\
 & channel dropout + SNR scaling & 79.27 & 5.60 & $6.32\cdot10^{-3}$ \\
 & signal drift + band noise & 78.07 & 5.54 & $2.37\cdot10^{-3}$ \\
 & channel dropout + signal drift & 78.52 & 6.43 & $2.21\cdot10^{-3}$ \\ \midrule
\multirow{4}{*}{DeepConvNet} & Baseline & 70.81 & 16.11 & - \\
 & phase Swap + band noise & 75.69 & 7.85 & $3.53\cdot10^{-2}$  \\
 & channel dropout + band noise & 75.48 & 8.12 & $3.27\cdot10^{-2}$ \\
 & channel dropout + phase swap & 74.99 & 7.41 & $3.19\cdot10^{-2}$ \\ \midrule
\multirow{4}{*}{EEGResNet} & Baseline & 72.94 & 7.81 & - \\
 & randome phase + signal drift & 74.18 & 6.26 & $3.37\cdot10^{-3}$ \\
 & phase swap + SNR scaling & 73.92 & 6.44 & $2.33\cdot10^{-3}$ \\
 & time reverse & 73.93 & 6.95 & $1.48\cdot10^{-3}$ \\ \bottomrule
\end{tabular}
\end{table*}

To ensure a fair comparison between TransformEEG and the other selected models, an optimal data augmentation composition was identified for each model.
Choosing the optimal data augmentation involved testing performance variations across all 100 possible augmentation compositions, as described in subsection 2.4.2 (Data augmentation).
Due to the high computational cost of this analysis, a 10-outer, 5-inner N-LNSO cross-validation was performed for each combination.
This approach halves the number of training instances. 
It also alters the baseline reference for the comparison.
To account for this change, a new baseline, derived from the same cross-validation setting, was included in the analysis to identify the best augmentation for each model.
Table \ref{tab: da selection} reports the top three data augmentations for each model, along with the corresponding new baseline reference.
TransformEEG results are not included in this section, as they are reported in subsection 3.2, Figure 3, Panels A and B of the paper.
The selection criteria follow those described in subsection 2.4.2 of the paper, which are restated here for clarity.

Determining the optimal data augmentation is a challenging process, as both the median (central measure) and the interquartile range (variability) are crucial in the comparison.
In addition, these two values evolve on different scales, increasing the challenges in weighting their improvement over the reference baseline.
In order to identify a consistent and objective measure, which keeps into account the relative improvements of both the baseline median and interquartile range, this work proposes the Augmentation Relative Improvement Score (ARIS).
ARIS is defined as follows

\begin{equation*}
    \textit{ARIS} = 
    \begin{cases}
        \begin{aligned}
        &0 \quad\text{if } (M_b>M)  \lor (\text{IQR}_b<\text{IQR}) \\
         &\frac{M_b-M}{M_b} \times \frac{\text{IQR}-\text{IQR}_b}{\text{IQR}_b}  \quad\text{otherwise}            
        \end{aligned}
    \end{cases}
\end{equation*}

where:
\WarningsOff
\begin{enumerate}[-]
    \item $M_b$ is the median baseline without data augmentation.
    \item $M$ is the current median value with data augmentation.
    \item $\text{IQR}_b$ is the baseline interquartile range.
    \item $\text{IQR}$ is the current interquartile range.
\end{enumerate}
\WarningsOn

This score ensures that data augmentations that do not improve both the median and interquartile range of the balanced accuracy are discarded.
Additionally, it ensures improvements are properly weighted by the scale of the baseline value.

\subsection{TransformEEG variants}
\label{subsec: transformeeg tuning}

This section presents the results of various TransformEEG variants used to quantitatively address certain model design choices.
First, it demonstrates that adding a positional embedding or a class token does not improve TransformEEG's performance, particularly in terms of result variability.
Next, it examines the effect of increasing the number of attention heads in the transformer encoder layers.

\subsubsection{Positional Encoding and Class Token}
\label{subsubsec: pos and cls}

\begin{figure}[!b]
    \centering
    \includegraphics[width=\linewidth]{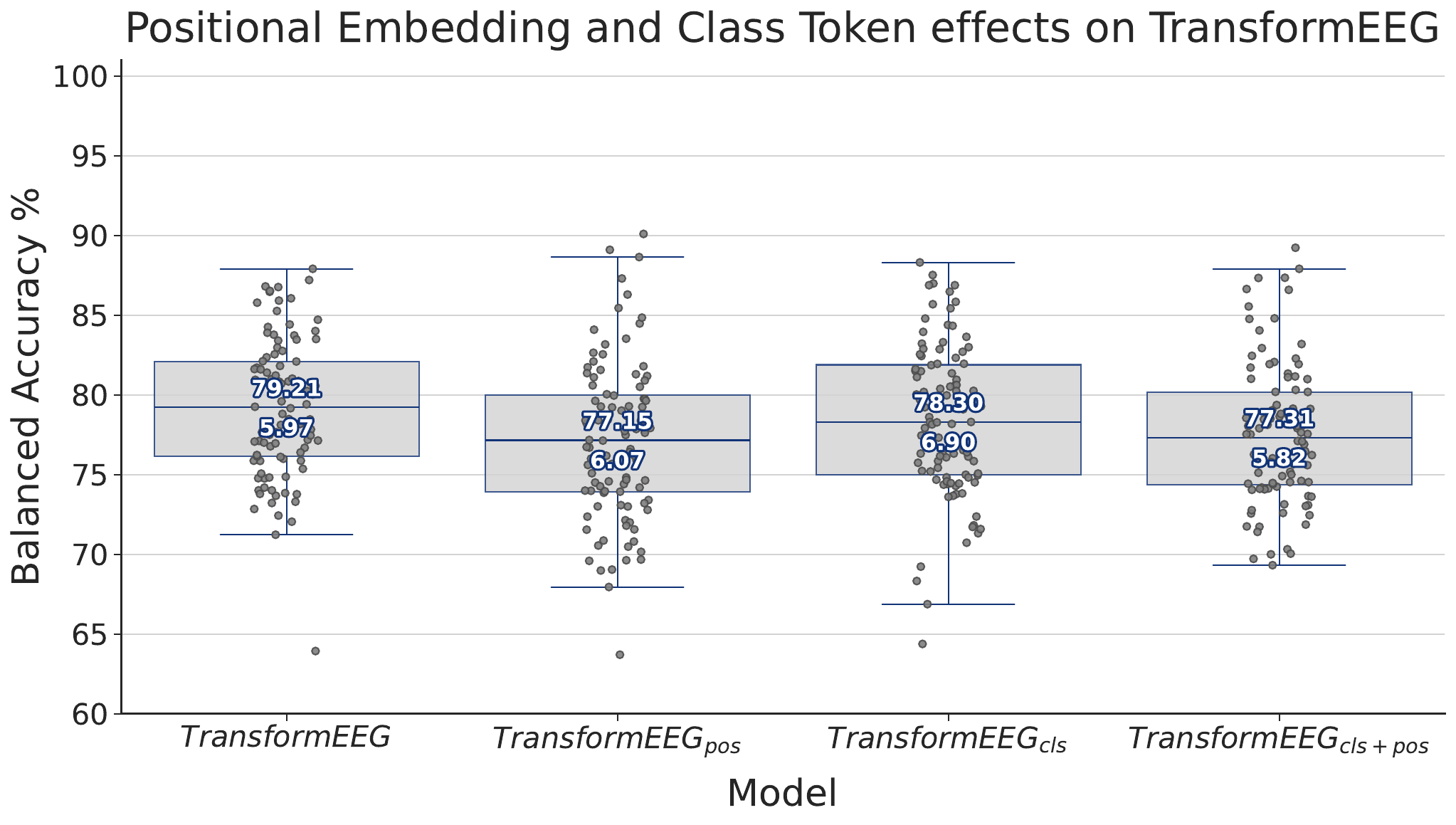}
    \caption{
    Comparison between TransformEEG and its variants that incorporate positional embedding, class token, or both.
    }
    \label{fig: pos and cls}
\end{figure}

Figure \ref{fig: pos and cls} shows the results of variants of TransformEEG that incorporate a positional embedding, a class token, or both, into the transformer encoder module.
All models are evaluated against a reference training pipeline that includes the optimal identified data augmentation masking plus time reverse.
As observed, none of the tested configurations improve upon the reference values.
Using both a class token and a positional embedding slightly decreases the IQR by 0.1\% but this decrease is accompanied by a 1.9\% drop in the median value.
These results demonstrate that adding these elements to the network does not provide a significant benefit.

\subsubsection{Changing the number of heads}
\label{subsubsec: heads}

\begin{figure}[!b]
    \centering
    \includegraphics[width=\linewidth]{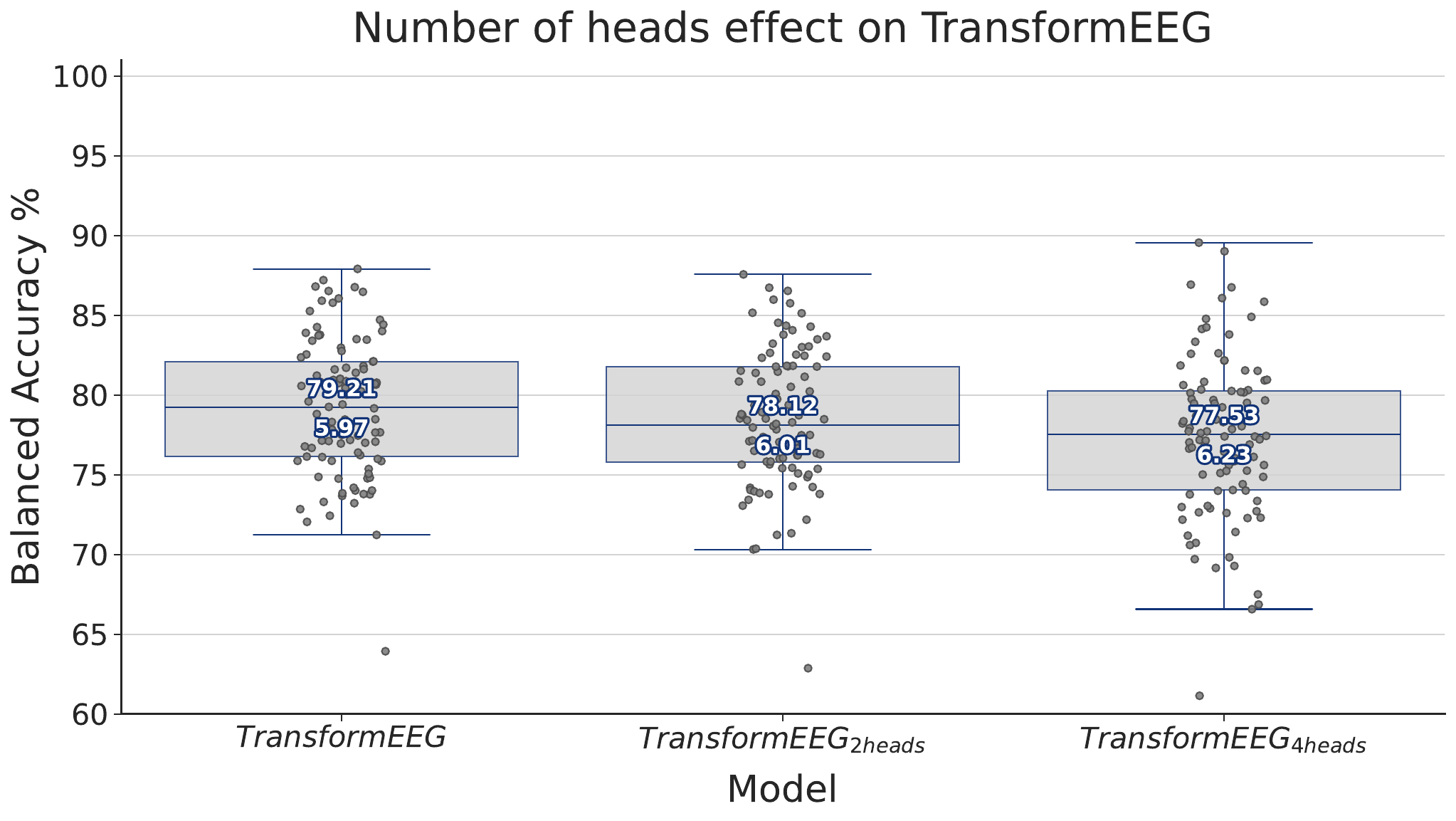}
    \caption{
    Comparison between TransformEEG and variants that use a transformer encoder layer with more heads. 
    }
    \label{fig: heads}
\end{figure}

Figure \ref{fig: heads} shows the results of TransformEEG variants that use a transformer encoder layer with more heads. 
The boxplots highlight that increasing the number of heads does not improve the performance of the model.
While using more attention heads allows TransformEEG to capture multiple types of relationships within the input data simultaneously, it also raises the risk of overfitting the training set, especially when the number of subjects is not very large.

\subsection{Windows length and overlap effects}
\label{subsec: window length and overlap}

\begin{table*}[h]
\caption{ShallowNet and transformEEG N-LNSO results with different window length and overlap}
\label{tab: win_and_over}
\begin{tabular}{ccccccccc}
    \toprule
    \multirow{3}{*}{\makecell{Window\\Length\\$[s]$}}
    & \multirow{3}{*}{\makecell{Window\\ Overlap\\\%}}
    & \multirow{3}{*}{\makecell{Number\\ of\\Samples}}
    &
    & \multicolumn{5}{c}{Balanced Accuracy} 
    \\ \cline{5-9}
    & & & & \multicolumn{2}{c}{transformEEG} & & \multicolumn{2}{c}{ShallowNet} 
    \\ \cline{5-6} \cline{8-9}
    & & & & Median & IQR & & Median & IQR
    \\ \midrule
    \multirow{4}{*}{2}
     &  0 & \num{44977}  & & 73.23 & 5.24 & & 76.20 & 9.78 \\
     & 25 & \num{59927}  & & 72.85 & 3.99 & & 76.84 & 9.71 \\
     & 50 & \num{90056}  & & 73.44 & 4.97 & & 76.13 & 8.10\\
     & 75 & \num{180893} & & 73.36 & 4.06 & & 77.29 & 9.88 \\
    \midrule 
    \multirow{4}{*}{4}
     &  0 & \num{22436} & & 74.66 & 4.74 & & 77.35 & 10.95 \\
     & 25 & \num{29939} & & 74.43 & 5.31 & & 78.57 & 11.35 \\
     & 50 & \num{44969} & & 74.11 & 5.39 & & 79.02 & 10.91 \\
     & 75 & \num{89376} & & 74.62 & 5.92 & & 79.32 & 9.87 \\
    \midrule 
    \multirow{4}{*}{8}
     &  0 & \num{11104} & & 75.36 & 5.52 & & 77.41 & 8.90 \\
     & 25 & \num{14939} & & 75.02 & 6.52 & & 78.23 & 9.66\\
     & 50 & \num{22311} & & 75.31 & 6.72 & & 77.68 & 10.27\\
     & 75 & \num{44277} & & 76.32 & 6.26 & & 79.23 & 11.74\\
    \midrule 
    \multirow{4}{*}{16}
     &  0 & \num{5426}  & & 75.47 & 7.26 & & 77.24 & 9.77\\
     & 25 & \num{7519}  & & 77.12 & 5.62 & & 77.90 & 10.77\\
     & 50 & \num{11102} & & 76.49 & 7.05 & & 78.38 & 10.58\\
     & 75 & \num{21736} & & 76.62 & 6.65 & & 78.94 & 10.02\\
    \midrule 
    \multirow{4}{*}{32}
     &  0 & \num{2606}  & & 75.38 & 8.40 & & 78.52 & 11.74 \\
     & 25 & \num{3730}  & & 76.05 & 7.56 & & 78.12 & 12.99 \\
     & 50 & \num{5422}  & & 76.34 & 7.65 & & 78.72 & 12.39 \\
     & 75 & \num{10404} & & 77.86 & 7.94 & & 80.48 & 12.68 \\
    \bottomrule 
\end{tabular}
\end{table*}

This section provides a supplementary analysis of the effects of two data partition hyperparameters: the window length and the window overlap.
These hyperparameters are crucial in determining how the same amount information is used to train and evaluate the performance of a deep neural network.

Increasing the window length allows for longer segments of the EEG signal to be considered during a forward pass.
This can help capture long-range patterns that might enhance the generalizability of an EEG deep learning model and, ultimately, its accuracy.
Additionally, longer windows better represent the overall spectral properties of the signal, ensuring that the activity registered in specific frequency bands of interest (e.g., theta, alpha, beta) is more reflective of the entire EEG signal.
However, extracting longer windows reduces the number of samples for training the model and increases GPU memory allocation.
Furthermore, the model must be designed to capture and exploite long-term patterns.
For example, in convolutional deep learning models from the EEGNet family, the usage of longer windows (if not handled by large pooling operations or strided convolutions) can lead to an unnecessary increase in model size and potentially unbalance the distribution of the weights across the network, with most parameters being concentrated on the final dense layer.
Consequently, selecting the appropriate window length and overlap is essential to reduce overfitting tendencies (resulting from too few samples) while maintaining effective learning capabilities (which may be compromised by excessively long windows).

Table \ref{tab: win_and_over} summarizes the results of multiple 10-outer, 10-inner N-LNSO cross-validation runs for transformEEG and ShallowNet models.
These runs use window lengths of 2, 4, 8, 16, and 32 seconds, with window overlaps of 0\%, 25\%, 50\%, and 75\%.
Each N-LNSO included a time reverse plus phase swap data augmentation composition to mitigate the reduction in dataset size associated with longer window lengths, without altering the power spectral density of the signals.
The median balanced accuracy increases with longer window lengths, particularly for the TransformEEG architecture.
This highlights the model's ability to capture long-range patterns within the EEG window.
However, the interquartile range also tends to increase, indicating greater variability in the results.
In contrast, the median balanced accuracy of ShallowNet appears primarily influenced by the overlap percentage.
Notably, this overlap does not significantly affect the variability of the results, which remain high regardless of the partition configuration.

From this analysis, it was determined that a window length of 16 seconds and an overlap of 25\% provides a good compromise between the median balanced accuracy and the interquartile range.
The usage of a higher overlap percentage was avoided to not create too similar samples that might penalize model generalizability, which is the key metric investigated in this study.

\subsection{Other strategies to improve model generalizability}
\label{subsec: other attempts}

This section presents the results of various efforts to improve the generalizability of TransformEEG. 
First, it provides a preliminary analysis of window aggregation techniques.
Next, it examines how the power spectral density of an EEG window can be incorporated into the TransformEEG architecture.
Finally, it reports on an attempt to integrate Common Spatial Pattern (CSP) during model training as an additional form of data scaling.

\subsubsection{Aggregating EEG windows predictions}
\label{subsubsec: aggregation}

\begin{figure*}[!t]
    \centering
    \includegraphics[width=0.95\textwidth]{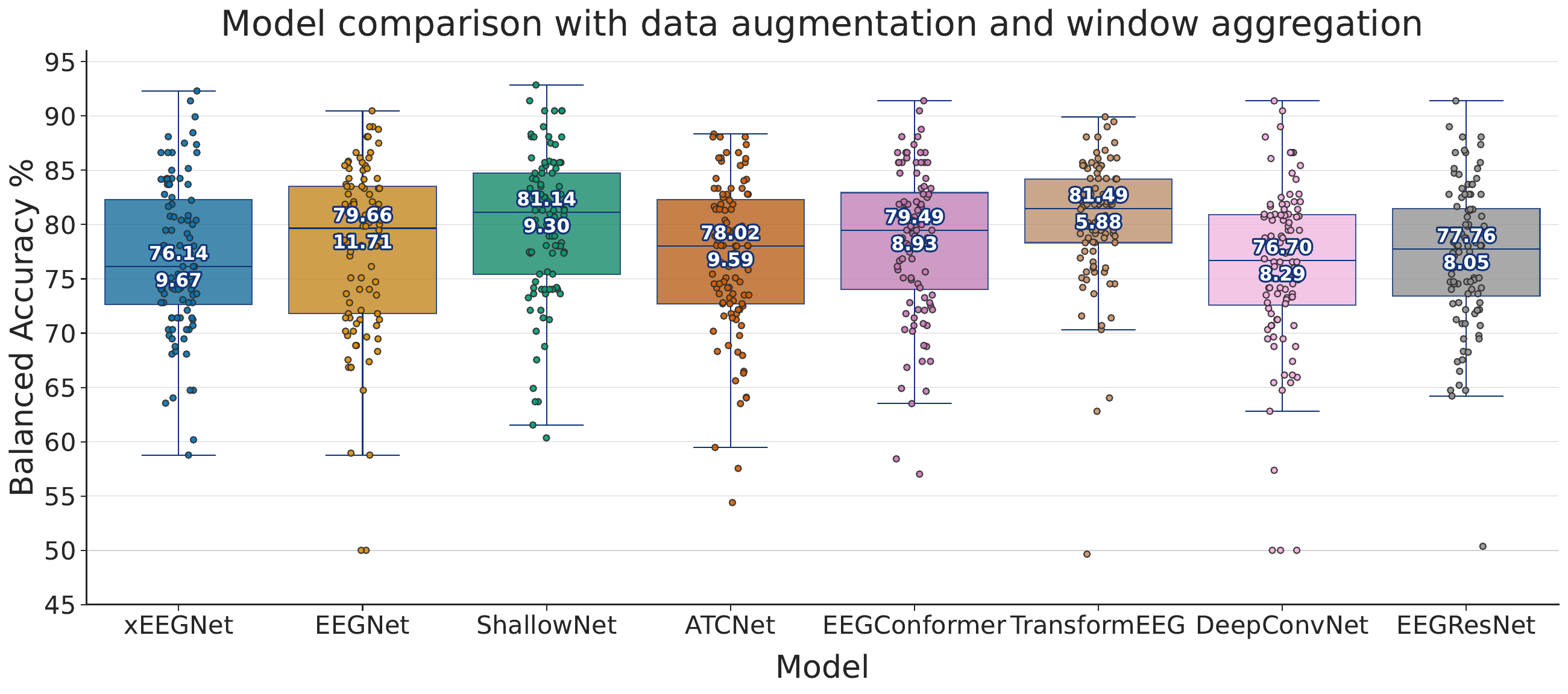}
    \caption{
    Performance comparison of selected models using a 10-outer, 10-inner N-LNSO cross-validation scheme.
    Models are arranged from left to right in order of increasing learnable parameters.
    All models were trained with their optimal data augmentation methods and evaluated through the aggregation procedure.
    }
    \label{fig: window_aggregation}
\end{figure*}

As discussed in Section 4 of the paper, the deep learning models investigated in this study take as input an EEG window in the time domain and output the probability that the window belongs to one of the investigated diseases.
While this approach aligns with optimal engineering solutions for addressing small dataset sizes and GPU memory limitations, it may not be the most effective method for disease detection in a clinical setting.
In pathology classification tasks, it is often more clinically relevant to determine whether the entire EEG originates from an individual with the target disease, rather than focusing solely on a single time window.
Figure \ref{fig: window_aggregation} illustrates the results when predictions from all the EEG windows are aggregated to produce a single classification for the entire signal.

Predictions are computed as follows:

\WarningsOff
\begin{enumerate}[\textbullet]
    \item Predictions are made for each EEG window using a standard classification threshold of 0.5. 
    \item A minimal ratio of positive windows is calculated using the EEG data from the validation set. This ratio represents the minimum percentage of windows that must be predicted as positive for the entire EEG to be classified as positive.
    \item Based on this ratio, a class label is assigned to each EEG in the test set, and the corresponding confusion matrix is computed.
\end{enumerate}
\WarningsOn

Overall, the median accuracy tends to improve compared to the training pipeline with data augmentation and threshold correction at the window level.
However, variability across models remains similar to that reported in the paper for the data augmentation plus threshold correction pipeline, although more outliers are observed.
It is also important to note that the proposed aggregation procedure alters the reference for model comparisons, as the confusion matrices are computed on fewer samples and thus give more weight to each prediction error.

\subsubsection{Addition of the PSD}
\label{subsubsec: psd}

\begin{figure*}[!t]
    \centering
    \includegraphics[width=0.7\linewidth]{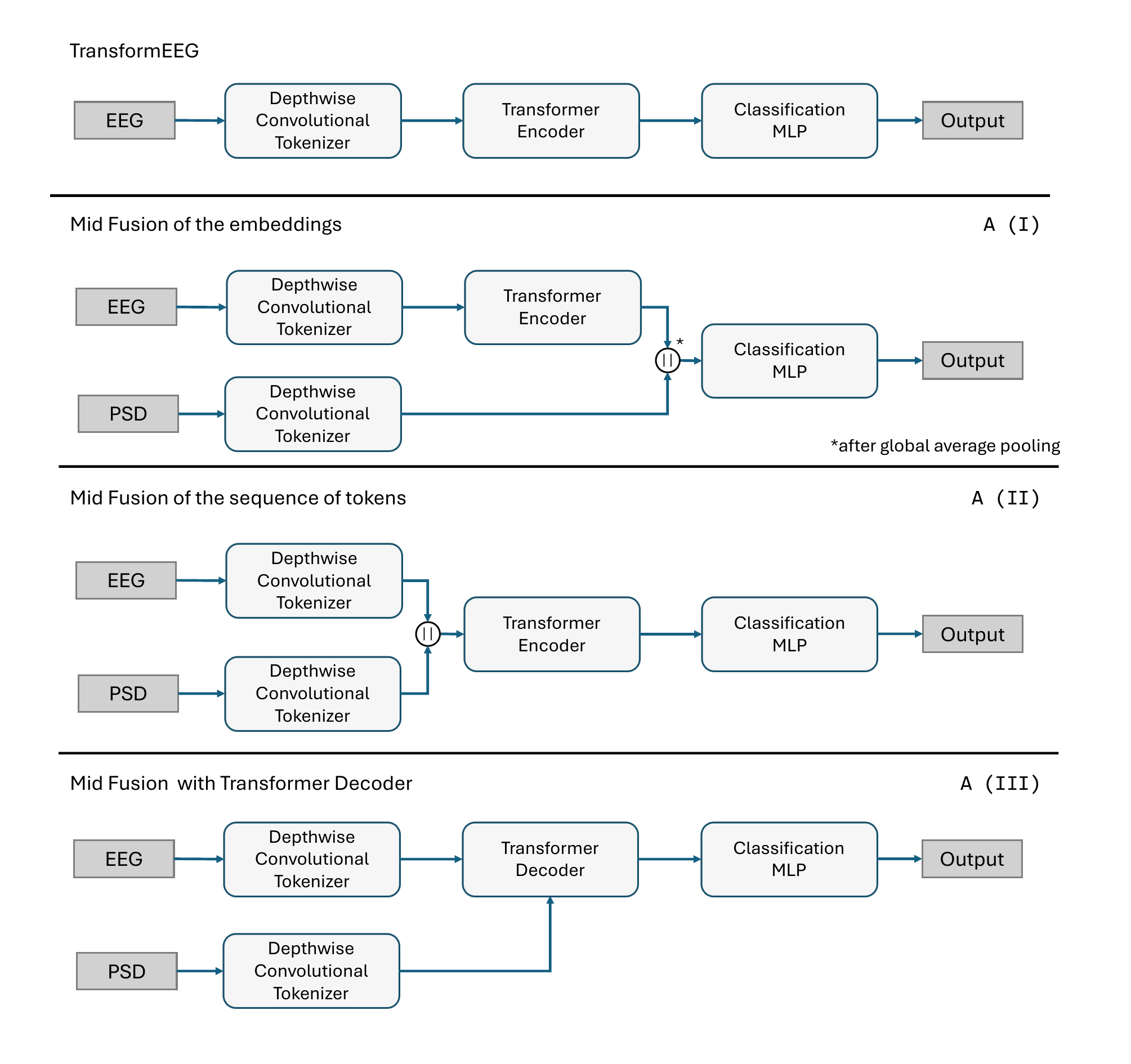}
    
    \includegraphics[width=0.7\linewidth]{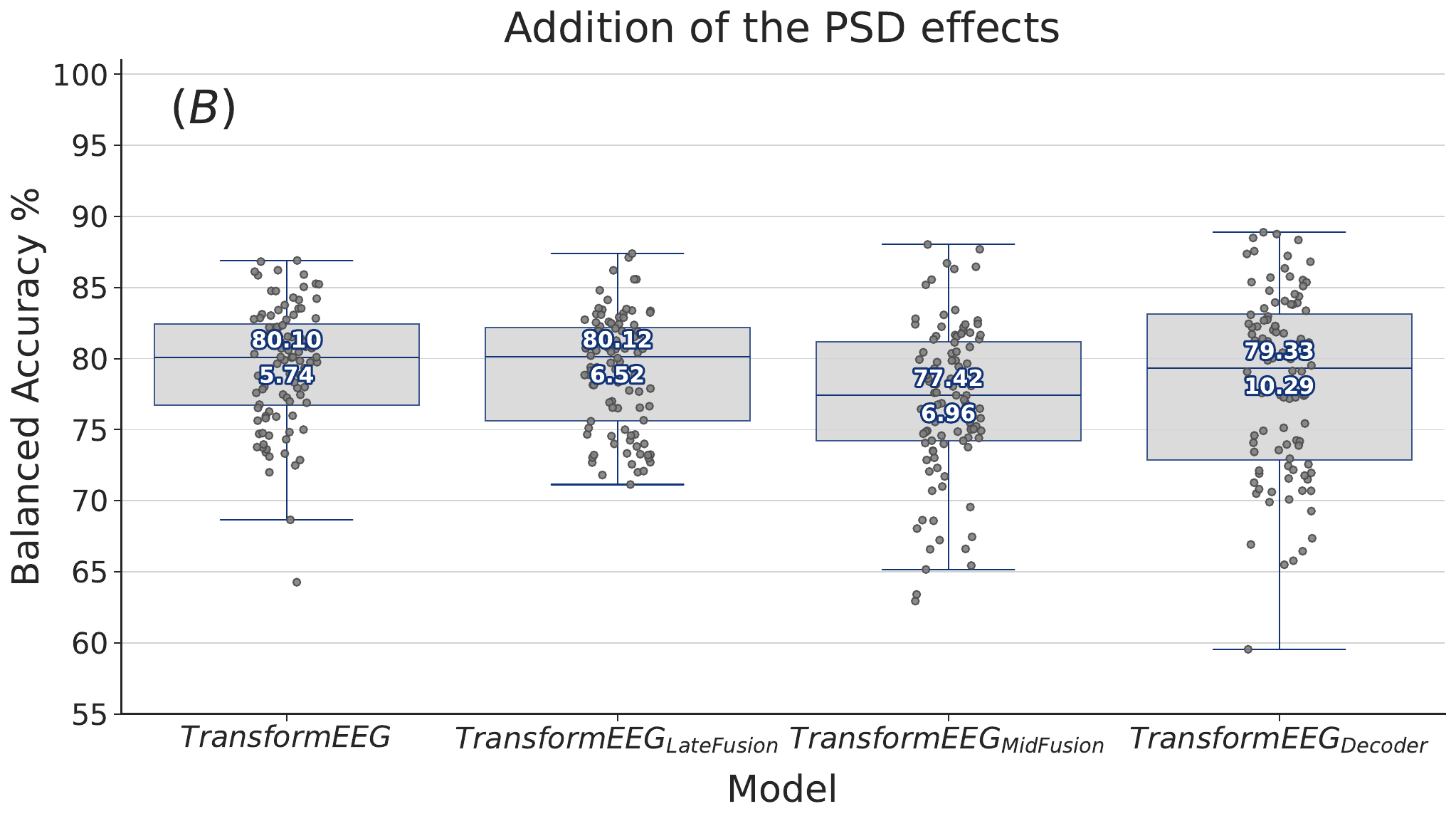}
    \caption{
    Effects of the addition of the PSD on the performance of TransformEEG.
    Panel A schematizes mid- and late-fusion strategies.
    Panel B shows the comparison between TransformEEG and variants that include the PSD branch.
    All models are trained with the masking plus flip horizontal data augmentation and threshold correction.
    }
    \label{fig: psd}
\end{figure*}

As discussed in Section 4 of the paper, TransformEEG does not process the EEG window in the time domain to specifically extract features that are linearly proportional to the mean power of EEG bands across channels.
While this design choice enhances the model's performance, it would be of great clinical interest to investigate whether combining both temporal and spectral information could further improve the model's generalizability.

This Section provides a preliminary analysis of how the PSD of an EEG window can be integrated within the transformEEG architecture and how this affects accuracy.
Figure \ref{fig: psd} illustrates three different PSD fusion strategies.
First, the PSD of every EEG window is extracted with the multitaper power spectral density estimate.
Next, a separate depthwise convolutional tokenizer creates channel-specific PSD tokens.
These representations are integrated with three different mid-fusion strategies.
The first mid-fusion strategy involves performing a global average pooling on the PSD tokens and concatenating the resulting vector with the temporal embedding (see Panel A.I).
The second strategy is based on the concatenation of PSD and temporal tokens before the transformer encoder (see Panel A.II).
The third mid-fusion strategy combines PSD and temporal tokens within a transformer decoder layer that performs cross-attention operations (see Panel A.III).
Regardless of the fusion strategy, incorporating spectral information through the PSD does not significantly improve the model's performance (see Panel B).

\subsubsection{Addition of the Common Spatial Pattern}
\label{subsubsec: csp}

This section presents the results of an unsuccessful attempt to incorporate Common Spatial pattern (CSP) as an additional data scaling method during model training.
CSP is a signal processing technique that transforms EEG data from the time domain into the spatial domain using a spatial filter.
The goal of this filter is to better differentiate between two classes by increasing the variance for one class (Parkinson's patients) and decreasing it for the other (healthy controls) within the same domain.
CSP is treated as a scaler and is computed using linear algebra and multivariate statistical methods (see Aljalal et al., 2022). During training, CSP filters are derived from the training data, with the first filter maximizing variance for the PD class and the last maximizing variance for healthy controls. Validation and test data are only transformed using these filters and are not involved in the CSP filter computation.

\begin{figure*}[!t]
    \centering
    \includegraphics[width=\textwidth]{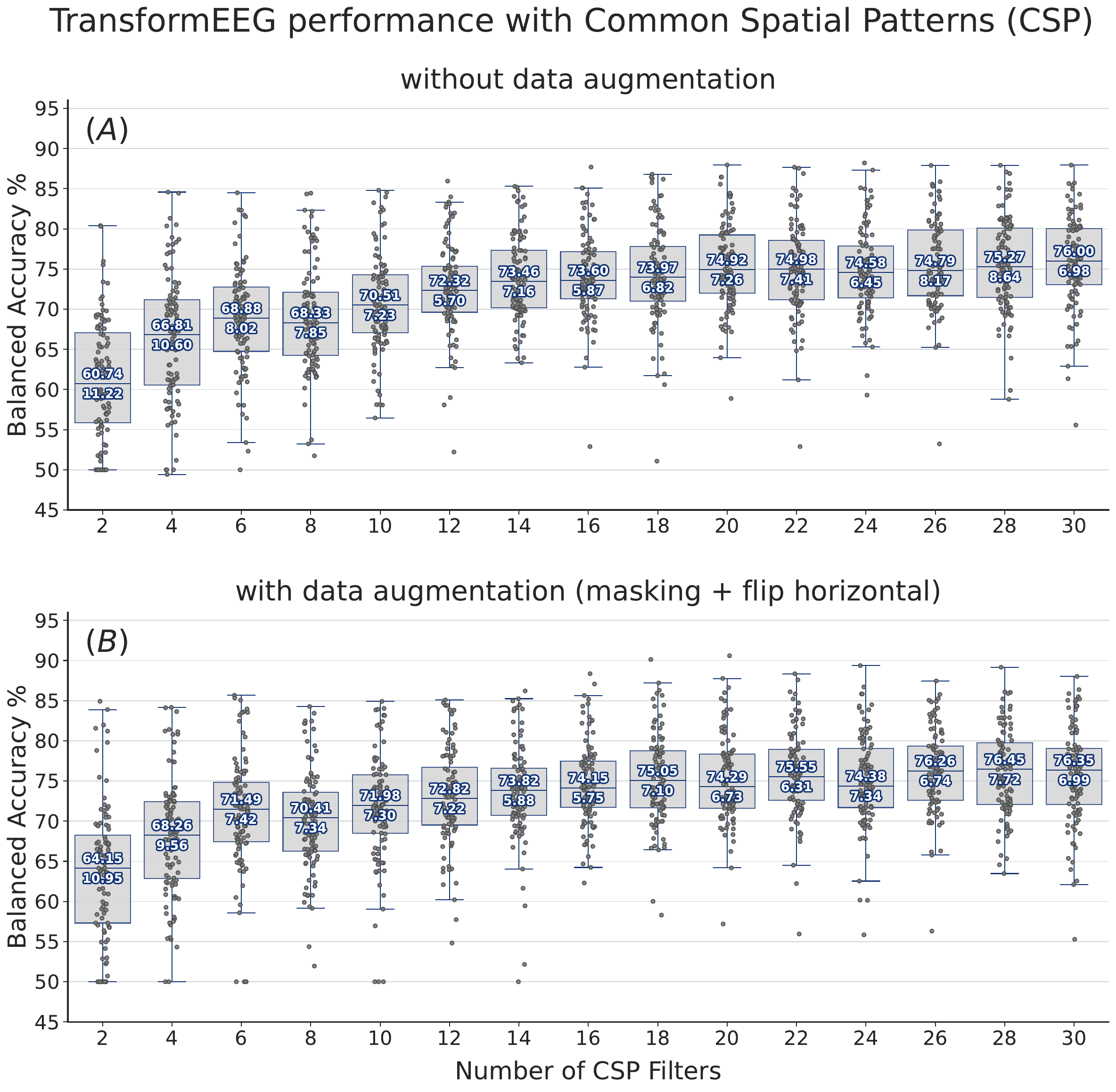}
    \caption{
    TransformEEG balanced accuracy variation when trained with an increasing number of CSP filters. Panel A shows the balanced accuracy using a baseline pipeline without data augmentation and threshold correction.
    Panel b shows the balanced accuracy using a baseline pipeline with data augmentation.
    }
    \label{fig: csp}
\end{figure*}

Figure \ref{fig: csp} illustrates the results of TransformEEG trained with an increasing number of CSP filters. Both baseline (Panel A) and data augmentation (Panel B) training settings are shown.
In both scenarios, adding CSP filters does not improve the model's performance.

\clearpage
\subsection{Additional Details on Model architectures}

This section offers detailed information about the deep learning models selected for this study.
A summary table is provided for each model, as indicated below: Table {\ref{tab:summaryEEGNet}} for EEGNet, Table {\ref{tab:summaryDeepConvNet}} for DeepConvNet, Table {\ref{tab:summaryShallow}} for ShallowConvNet, Table {\ref{tab:summaryeegconformer}} for EEGConformer, Table 
{\ref{tab:summaryXeegnet}} for xEEGNet, Table {\ref{tab:summaryatcnet}} for ATCNet, Table {\ref{tab:summaryTResNer}} for EEGResNet, Table {\ref{tab:summarytransformeeg}} for TransformEEG.
Each table outlines important characteristics, including input shape, output shape, kernel size, number of convolutional groups, and the number of parameters for each layer in the network. 
Additionally, activation functions and dimensional operations (such as flattening and unsqueezing) are included for clarity.
The input and output sizes correspond to the binary Parkinson's Disease detection task.
Specifically, the models receives an EEG trial window consisting of 32 channels and 2000 samples (equivalent to 16 seconds at 125 Hz) and output the probability of such a sample being from a PD recording.
The variable N denotes the batch size, which was set to 64 for this study, as explained in subsection 2.4.3.
A complete implementation of each deep learning model is accessible in the openly available SelfEEG library's code base (version 0.2.0+) or in the GitHub repository associated with this study.
\clearpage
\texttt{
\begin{table*}[bp!]
    \caption{Summary Table of EEGNet.}
    \begin{tabular}{llllll}
        \toprule 
        Layer(type) & InputShape & OutputShape & KernelShape & Groups & Param\# \\ 
        \midrule 
        EEGNet & $[\text{N},32,2000]$ & $[\text{N},1]$ & -- & -- & -- \\ 
        +EEGNet Encoder & $[\text{N},32,2000]$ & $[\text{N},992]$ & -- & -- & -- \\
        |\quad+Unsqueeze & $[\text{N},32,2000]$ & $[\text{N},1,32,2000]$ & -- & -- & -- \\ 
        |\quad+Conv2d & $[\text{N},1,32,2000]$ & $[\text{N},8,32,2000]$ & $[1,64]$ & 1 & 512 \\ 
        |\quad+BatchNorm2d & $[\text{N},8,32,2000]$ & $[\text{N},8,32,2000]$ & -- & -- & 16 \\ 
        |\quad+Conv2d & $[\text{N},8,32,2000]$ & $[\text{N},16,1,2000]$ & $[32,1]$ & 8 & 512 \\ 
        |\quad+BatchNorm2d & $[\text{N},16,1,2000]$ & $[\text{N},16,1,2000]$ & -- & -- & 32 \\ 
        |\quad+ELU & $[\text{N},16,1,2000]$ & $[\text{N},16,1,2000]$ & -- & -- & -- \\ 
        |\quad+AvgPool2d & $[\text{N},16,1,2000]$ & $[\text{N},16,1,500]$ & $[1,4]$ & -- & -- \\ 
        |\quad+Dropout & $[\text{N},16,1,500]$ & $[\text{N},16,1,500]$ & -- & -- & -- \\ 
        |\quad+SeparableConv2d & $[\text{N},16,1,500]$ & $[\text{N},16,1,500]$ & -- & -- & -- \\ 
        |\quad|\quad+Conv2d & $[\text{N},16,1,500]$ & $[\text{N},16,1,500]$ & $[1,16]$ & 16 & 256 \\ 
        |\quad|\quad+Conv2d & $[\text{N},16,1,500]$ & $[\text{N},16,1,500]$ & $[1,1]$ & 1 & 256 \\ 
        |\quad+BatchNorm2d & $[\text{N},16,1,500]$ & $[\text{N},16,1,500]$ & -- & -- & 32 \\ 
        |\quad+ELU & $[\text{N},16,1,500]$ & $[\text{N},16,1,500]$ & -- & -- & -- \\ 
        |\quad+AvgPool2d & $[\text{N},16,1,500]$ & $[\text{N},16,1,62]$ & $[1,8]$ & -- & -- \\ 
        |\quad+Dropout & $[\text{N},16,1,62]$ & $[\text{N},16,1,62]$ & -- & -- & -- \\ 
        |\quad+Flatten & $[\text{N},16,1,62]$ & $[\text{N},992]$ & -- & -- & -- \\ 
        +Dense & $[\text{N},992]$ & $[\text{N},1]$ & -- & -- & 993 \\ 
        \bottomrule  
    \end{tabular}
    \label{tab:summaryEEGNet}
\end{table*} 
}

\texttt{
\begin{table*}
    \centering
    \caption{Summary Table of DeepConvNet.}
    \begin{tabular}{llllll}
        \toprule 
        Layer(type) & InputShape & OutputShape & KernelShape & Groups & Param\# \\ 
        \midrule 
        DeepConvNet & $[\text{N},32,2000]$ & $[\text{N},1]$ & -- & -- & -- \\ 
        +DeepConvNet Encoder & $[\text{N},32,2000]$ & $[\text{N},4000]$ & -- & -- & -- \\
        |\quad+Unsqueeze & $[\text{N},32,2000]$ & $[\text{N},1,32,2000]$ & -- & -- & -- \\ 
        |\quad+Conv2d & $[\text{N},1,32,2000]$ & $[\text{N},25,32,1991]$ & $[1,10]$ & 1 & 275 \\ 
        |\quad+Conv2d & $[\text{N},25,32,1991]$ & $[\text{N},25,1,1991]$ & $[32,1]$ & 1 & 20,025 \\ 
        |\quad+BatchNorm2d & $[\text{N},25,1,1991]$ & $[\text{N},25,1,1991]$ & -- & -- & 50 \\ 
        |\quad+ELU & $[\text{N},25,1,1991]$ & $[\text{N},25,1,1991]$ & -- & -- & -- \\ 
        |\quad+MaxPool2d & $[\text{N},25,1,1991]$ & $[\text{N},25,1,663]$ & $[1,3]$ & -- & -- \\ 
        |\quad+Dropout & $[\text{N},25,1,663]$ & $[\text{N},25,1,663]$ & -- & -- & -- \\ 
        |\quad+Conv2d & $[\text{N},25,1,663]$ & $[\text{N},50,1,654]$ & $[1,10]$ & 1 & 12,550 \\ 
        |\quad+BatchNorm2d & $[\text{N},50,1,654]$ & $[\text{N},50,1,654]$ & -- & -- & 100 \\ 
        |\quad+ELU & $[\text{N},50,1,654]$ & $[\text{N},50,1,654]$ & -- & -- & -- \\ 
        |\quad+MaxPool2d & $[\text{N},50,1,654]$ & $[\text{N},50,1,218]$ & $[1,3]$ & -- & -- \\ 
        |\quad+Dropout & $[\text{N},50,1,218]$ & $[\text{N},50,1,218]$ & -- & -- & -- \\ 
        |\quad+Conv2d & $[\text{N},50,1,218]$ & $[\text{N},100,1,209]$ & $[1,10]$ & 1 & 50,100 \\ 
        |\quad+BatchNorm2d & $[\text{N},100,1,209]$ & $[\text{N},100,1,209]$ & -- & -- & 200 \\ 
        |\quad+ELU & $[\text{N},100,1,209]$ & $[\text{N},100,1,209]$ & -- & -- & -- \\ 
        |\quad+MaxPool2d & $[\text{N},100,1,209]$ & $[\text{N},100,1,69]$ & $[1,3]$ & -- & -- \\ 
        |\quad+Dropout & $[\text{N},100,1,69]$ & $[\text{N},100,1,69]$ & -- & -- & -- \\ 
        |\quad+Conv2d & $[\text{N},100,1,69]$ & $[\text{N},200,1,60]$ & $[1,10]$ & 1 & 200,200 \\ 
        |\quad+BatchNorm2d & $[\text{N},200,1,60]$ & $[\text{N},200,1,60]$ & -- & -- & 400 \\ 
        |\quad+ELU & $[\text{N},200,1,60]$ & $[\text{N},200,1,60]$ & -- & -- & -- \\ 
        |\quad+MaxPool2d & $[\text{N},200,1,60]$ & $[\text{N},200,1,20]$ & $[1,3]$ & -- & -- \\ 
        |\quad+Dropout & $[\text{N},200,1,20]$ & $[\text{N},200,1,20]$ & -- & -- & -- \\ 
        |\quad+Flatten & $[\text{N},200,1,20]$ & $[\text{N},4000]$ & -- & -- & -- \\ 
        +Dense & $[\text{N},4000]$ & $[\text{N},1]$ & -- & -- & 4,001 \\ 
        \bottomrule 
    \end{tabular}
    \label{tab:summaryDeepConvNet}
\end{table*}
}

\texttt{
\begin{table*}
    \centering
    \caption{Summary Table of ShallowNet.}
    \begin{tabular}{llllll}
        \toprule 
        Layer(type) & Input Shape & Output Shape & Kernel Shape & Groups & Param\# \\ 
        \midrule 
        ShallowNet & $[\text{N},32,2000]$ & $[\text{N},1]$ & -- & -- & -- \\ 
        +ShallowNet Encoder & $[\text{N},32,2000]$ & $[\text{N},5080]$ & -- & -- & -- \\ 
        |\quad+Unsqueeze & $[\text{N},32,2000]$ & $[\text{N},1,32,2000]$ & -- & -- & -- \\ 
        |\quad+Conv2d & $[\text{N},1,32,2000]$ & $[\text{N},40,32,1976]$ & $[1,25]$ & 1 & 1,040 \\ 
        |\quad+Conv2d & $[\text{N},40,32,1976]$ & $[\text{N},40,1,1976]$ & $[32,1]$ & 1 & 51,240 \\ 
        |\quad+BatchNorm2d & $[\text{N},40,1,1976]$ & $[\text{N},40,1,1976]$ & -- & -- & 80 \\ 
        |\quad+Square & $[\text{N},40,1,1976]$ & $[\text{N},40,1,1976]$ & -- & -- & -- \\ 
        |\quad+AvgPool2d & $[\text{N},40,1,1976]$ & $[\text{N},40,1,127]$ & $[1,75]$ & -- & -- \\  
        |\quad+Log & $[\text{N},40,1,27]$ & $[\text{N},40,1,27]$ & -- & -- & -- \\
        |\quad+Dropout & $[\text{N},40,1,127]$ & $[\text{N},40,1,127]$ & -- & -- & -- \\ 
        |\quad+Flatten & $[\text{N},40,1,127]$ & $[\text{N},5080]$ & -- & -- & -- \\ 
        +Linear & $[\text{N},5080]$ & $[\text{N},1]$ & -- & -- & 5,081 \\ 
        \bottomrule 
    \end{tabular}
    \label{tab:summaryShallow}
\end{table*}
}

\texttt{
\begin{table*}
    \centering
    \caption{Summary Table of EEGConformer.}
    \begin{tabular}{llllll}
        \toprule 
        Layer(type) & InputShape & OutputShape & KernelShape & Groups & Param\# \\ 
        \midrule 
        EEGConformer & $[\text{N},32,2000]$ & $[\text{N},1]$ & -- & -- & -- \\ 
        +Tokenizer & $[\text{N},32,2000]$ & $[\text{N},127,40]$ & -- & -- & -- \\
        |\quad+Unsqueeze & $[\text{N},32,2000]$ & $[\text{N},1,32,2000]$ & -- & -- & -- \\ 
        |\quad+Conv2d & $[\text{N},1,32,2000]$ & $[\text{N},40,32,1976]$ & $[1,25]$ & 1 & 1,040 \\ 
        |\quad+Conv2d & $[\text{N},40,32,1976]$ & $[\text{N},40,1,1976]$ & $[32,1]$ & 1 & 51,240 \\ 
        |\quad+BatchNorm2d & $[\text{N},40,1,1976]$ & $[\text{N},40,1,1976]$ & -- & -- & 80 \\ 
        |\quad+ELU & $[\text{N},40,1,1976]$ & $[\text{N},40,1,1976]$ & -- & -- & -- \\ 
        |\quad+AvgPool2d & $[\text{N},40,1,1976]$ & $[\text{N},40,1,127]$ & $[1,75]$ & -- & -- \\ 
        |\quad+Dropout & $[\text{N},40,1,127]$ & $[\text{N},40,1,127]$ & -- & -- & -- \\ 
        |\quad+Conv2d & $[\text{N},40,1,127]$ & $[\text{N},40,1,127]$ & $[1,1]$ & 1 & 1,640 \\ 
        |\quad+Rearrange & $[\text{N},40,1,127]$ & $[\text{N},127,40]$ & -- & -- & -- \\
        +TransformerEncoder &  $[\text{N},127,40]$ & $[\text{N},127,40]$ & -- & -- & -- \\ 
        |\quad+TransformerEncoderLayer & $[\text{N},127,40]$ & $[\text{N},127,40]$ & -- & -- & -- \\ 
        |\quad|\quad+MultiheadAttention & $[\text{N},127,40]$ & $[\text{N},127,40]$ & -- & -- & 6,560 \\ 
        |\quad|\quad+Dropout & $[\text{N},127,40]$ & $[\text{N},127,40]$ & -- & -- & -- \\ 
        |\quad|\quad+LayerNorm & $[\text{N},127,40]$ & $[\text{N},127,40]$ & -- & -- & 80 \\ 
        |\quad|\quad+Linear & $[\text{N},127,40]$ & $[\text{N},127,160]$ & -- & -- & 6,560 \\ 
        |\quad|\quad+Dropout & $[\text{N},127,160]$ & $[\text{N},127,160]$ & -- & -- & -- \\ 
        |\quad|\quad+Linear & $[\text{N},127,160]$ & $[\text{N},127,40]$ & -- & -- & 6,440 \\ 
        |\quad|\quad+Dropout & $[\text{N},127,40]$ & $[\text{N},127,40]$ & -- & -- & -- \\ 
        |\quad|\quad+LayerNorm & $[\text{N},127,40]$ & $[\text{N},127,40]$ & -- & -- & 80 \\
        |\quad+TransformerEncoderLayer & $[\text{N},127,40]$ & $[\text{N},127,40]$ & -- & -- & 19,720 \\
        |\quad+TransformerEncoderLayer & $[\text{N},127,40]$ & $[\text{N},127,40]$ & -- & -- & 19,720 \\
        |\quad+TransformerEncoderLayer & $[\text{N},127,40]$ & $[\text{N},127,40]$ & -- & -- & 19,720 \\
        |\quad+TransformerEncoderLayer & $[\text{N},127,40]$ & $[\text{N},127,40]$ & -- & -- & 19,720 \\
        |\quad+TransformerEncoderLayer & $[\text{N},127,40]$ & $[\text{N},127,40]$ & -- & -- & 19,720 \\
        +ClassificationHead & $[\text{N},127,40]$ & $[\text{N},1]$ & -- & -- & -- \\
        |\quad+Rearrange & $[\text{N},127,40]$ & $[\text{N},40,127]$ & -- & -- & -- \\
        |\quad+AdaptiveAvgPool2d & $[\text{N},40,127]$ & $[\text{N},40]$ & -- & -- & -- \\ 
        |\quad+LayerNorm & $[\text{N},40]$ & $[\text{N},40]$ & -- & -- & 80 \\ 
        |\quad+Linear & $[\text{N},40]$ & $[\text{N},256]$ & -- & -- & 10,496 \\ 
        |\quad+ELU & $[\text{N},256]$ & $[\text{N},256]$ & -- & -- & -- \\ 
        |\quad+Dropout & $[\text{N},256]$ & $[\text{N},256]$ & -- & -- & -- \\ 
        |\quad+Linear & $[\text{N},256]$ & $[\text{N},32]$ & -- & -- & 8,224 \\ 
        |\quad+ELU & $[\text{N},32]$ & $[\text{N},32]$ & -- & -- & -- \\ 
        |\quad+Dropout & $[\text{N},32]$ & $[\text{N},32]$ & -- & -- & -- \\ 
        |\quad+Linear & $[\text{N},32]$ & $[\text{N},1]$ & -- & -- & 33 \\ 
        \bottomrule  
    \end{tabular}
    \label{tab:summaryeegconformer}
\end{table*}
}

\texttt{
\begin{table*}
    \centering
    \caption{Summary Table of XEEGNet. Parameters between round brackets are frozen.}
    \begin{tabular}{llllll}
        \toprule 
        Layer(type) & Input Shape & Output Shape & Kernel Shape & Groups & Param\# \\ 
        \midrule 
        XEEGNet & $[\text{N},32,2000]$ & $[\text{N},1]$ & -- & -- & -- \\ 
        +XEEGNet Encoder & $[\text{N},32,2000]$ & $[\text{N},7]$ & -- & -- & -- \\ 
        |\quad+Unsqueeze & $[\text{N},32,2000]$ & $[\text{N},1,32,2000]$ & -- & -- & -- \\ 
        |\quad+Conv2d & $[\text{N},1,32,2000]$ & $[\text{N},7,32,1876]$ & $[1,125]$ & 1 & (875) \\ 
        |\quad+Conv2d & $[\text{N},7,32,1876]$ & $[\text{N},7,1,1876]$ & $[32,1]$ & 7 & 224 \\ 
        |\quad+BatchNorm2d & $[\text{N},7,1,1876]$ & $[\text{N},7,1,1876]$ & -- & -- & 14 \\ 
        |\quad+Square & $[\text{N},40,1,1976]$ & $[\text{N},40,1,1976]$ & -- & -- & -- \\ 
        |\quad+AdaptiveAvgPool2d & $[\text{N},7,1,1876]$ & $[\text{N},7,1,1]$ & -- & -- & -- \\   
        |\quad+Log & $[\text{N},40,1,27]$ & $[\text{N},40,1,27]$ & -- & -- & -- \\
        |\quad+Dropout & $[\text{N},7,1,1]$ & $[\text{N},7,1,1]$ & -- & -- & -- \\ 
        |\quad+Flatten & $[\text{N},7,1,1]$ & $[\text{N},7]$ & -- & -- & -- \\ 
        +Linear & $[\text{N},7]$ & $[\text{N},1]$ & -- & -- & 7 \\ 
        \bottomrule  
    \end{tabular}
    \label{tab:summaryXeegnet}
\end{table*}
}

\texttt{
\begin{table*}
    \centering
    \caption{Summary Table of ATCNet.}
    \begin{tabular}{llllll}
        \toprule 
        Layer(type) & InputShape & OutputShape & KernelShape & Groups & Param\# \\ 
        \midrule 
        ATCNet & $[\text{N},32,2000]$ & $[\text{N},1]$ & -- & -- & -- \\ 
        +Sequential & $[\text{N},1,32,2000]$ & $[\text{N},32,1,71]$ & -- & -- & -- \\ 
        |\quad+Conv2d & $[\text{N},1,32,2000]$ & $[\text{N},16,32,2000]$ & $[1,31]$ & 1 & 496 \\ 
        |\quad+BatchNorm2d & $[\text{N},16,32,2000]$ & $[\text{N},16,32,2000]$ & -- & -- & 32 \\ 
        |\quad+Conv2d & $[\text{N},16,32,2000]$ & $[\text{N},32,1,2000]$ & $[32,1]$ & 16 & 1,024 \\ 
        |\quad+BatchNorm2d & $[\text{N},32,1,2000]$ & $[\text{N},32,1,2000]$ & -- & -- & 64 \\ 
        |\quad+ELU & $[\text{N},32,1,2000]$ & $[\text{N},32,1,2000]$ & -- & -- & -- \\ 
        |\quad+AvgPool2d & $[\text{N},32,1,2000]$ & $[\text{N},32,1,500]$ & $[1,4]$ & -- & -- \\ 
        |\quad+Dropout & $[\text{N},32,1,500]$ & $[\text{N},32,1,500]$ & -- & -- & -- \\ 
        |\quad+Conv2d & $[\text{N},32,1,500]$ & $[\text{N},32,1,500]$ & $[1,16]$ & 1 & 16,384 \\ 
        |\quad+BatchNorm2d & $[\text{N},32,1,500]$ & $[\text{N},32,1,500]$ & -- & -- & 64 \\ 
        |\quad+ELU & $[\text{N},32,1,500]$ & $[\text{N},32,1,500]$ & -- & -- & -- \\ 
        |\quad+AdaptiveAvgPool2d & $[\text{N},32,1,500]$ & $[\text{N},32,1,71]$ & -- & -- & -- \\ 
        |\quad+Dropout & $[\text{N},32,1,71]$ & $[\text{N},32,1,71]$ & -- & -- & -- \\
        |\quad+Rearrange & $[\text{N},32,1,71]$ & $[\text{N},71,32]$ & -- & -- & -- \\
        +ATC Module (5 in parallel) & $[\text{N},71,32]$ & $[\text{N},1]$ & -- & -- & -- \\ 
        |\quad+ATCmha & $[\text{N},67,32]$ & $[\text{N},67,32]$ & -- & -- & -- \\ 
        |\quad|\quad+LayerNorm & $[\text{N},67,32]$ & $[\text{N},67,32]$ & -- & -- & 4,288 \\ 
        |\quad|\quad+MultiheadAttention & $[\text{N},67,32]$ & $[\text{N},67,32]$ & -- & -- & 4,224 \\ 
        |\quad|\quad+Dropout & $[\text{N},67,32]$ & $[\text{N},67,32]$ & -- & -- & -- \\ 
        |\quad+ATCtcn & $[\text{N},32,67]$ & $[\text{N},32,67]$ & -- & -- & -- \\ 
        |\quad|\quad+ATCtcn Block & $[\text{N},32,67]$ & $[\text{N},32,67]$ & -- & -- & -- \\ 
        |\quad|\quad|\quad+Conv1d & $[\text{N},32,67]$ & $[\text{N},32,67]$ & $[4]$ & 1 & 4,128 \\ 
        |\quad|\quad|\quad+BatchNorm1d & $[\text{N},32,67]$ & $[\text{N},32,67]$ & -- & -- & 64 \\ 
        |\quad|\quad|\quad+ELU & $[\text{N},32,67]$ & $[\text{N},32,67]$ & -- & -- & -- \\ 
        |\quad|\quad|\quad+Dropout & $[\text{N},32,67]$ & $[\text{N},32,67]$ & -- & -- & -- \\ 
        |\quad|\quad|\quad+Conv1d & $[\text{N},32,67]$ & $[\text{N},32,67]$ & $[4]$ & 1 & 4,128 \\ 
        |\quad|\quad|\quad+BatchNorm1d & $[\text{N},32,67]$ & $[\text{N},32,67]$ & -- & -- & 64 \\ 
        |\quad|\quad|\quad+ELU & $[\text{N},32,67]$ & $[\text{N},32,67]$ & -- & -- & -- \\ 
        |\quad|\quad|\quad+Dropout & $[\text{N},32,67]$ & $[\text{N},32,67]$ & -- & -- & -- \\ 
        |\quad|\quad+ATCtcn Block & $[\text{N},32,67]$ & $[\text{N},32,67]$ & -- & -- & 8,384 \\  
        |\quad+Linear & $[\text{N},32]$ & $[\text{N},1]$ & -- & -- & 33 \\
        +Stack & $[\text{N},1]$ & $[5,\text{N},1]$ & -- & -- & -- \\
        +Mean & $[5,\text{N},1]$ & $[\text{N},1]$ & -- & -- & -- \\ 
        \bottomrule 
    \end{tabular}
    \label{tab:summaryatcnet}
\end{table*}
}

\texttt{
\begin{table*}
    \centering
    \caption{Summary Table of EEGResNet. ResBlocks include a skip connection.}
    \begin{tabular}{llllll}
        \toprule 
        Layer(type) & InputShape & OutputShape & KernelShape & Groups & Param\# \\ 
        \midrule 
        EEGResNet & $[\text{N},32,2000]$ & $[\text{N},1]$ & -- & -- & -- \\ 
        +EEGResNet Encoder & $[\text{N},32,2000]$ & $[\text{N},256]$ & -- & -- & -- \\
        |\quad+Unsqueeze & $[\text{N},32,2000]$ & $[\text{N},1,32,2000]$ & -- & -- & -- \\ 
        |\quad+Conv2d & $[\text{N},1,32,2000]$ & $[\text{N},32,32,2000]$ & $[1,9]$ & 1 & 320 \\ 
        |\quad+BatchNorm2d & $[\text{N},32,32,2000]$ & $[\text{N},32,32,2000]$ & -- & -- & 64 \\ 
        |\quad+ReLU & $[\text{N},32,32,2000]$ & $[\text{N},32,32,2000]$ & -- & -- & -- \\ 
        +Stage 1 & $[\text{N},32,32,2000]$ & $[\text{N},32,32,1000]$ & -- & -- & -- \\ 
        |\quad+Temporal ResBlock (+ Downsampling) & $[\text{N},32,32,2000]$ & $[\text{N},32,32,1000]$ & -- & -- & -- \\ 
        |\quad|\quad+Downsampling & $[\text{N},32,32,2000]$ & $[\text{N},32,32,1000]$ & -- & -- & -- \\ 
        |\quad|\quad|\quad+Conv2d & $[\text{N},32,32,2000]$ & $[\text{N},32,32,1000]$ & $[1,1]$ & 1 & 1,024 \\ 
        |\quad|\quad|\quad+BatchNorm2d & $[\text{N},32,32,1000]$ & $[\text{N},32,32,1000]$ & -- & -- & 64 \\ 
        |\quad|\quad+Conv2d & $[\text{N},32,32,2000]$ & $[\text{N},32,32,1000]$ & $[1,9]$ & 1 & 9,216 \\ 
        |\quad|\quad+BatchNorm2d & $[\text{N},32,32,1000]$ & $[\text{N},32,32,1000]$ & -- & -- & 64 \\ 
        |\quad|\quad+ReLU & $[\text{N},32,32,1000]$ & $[\text{N},32,32,1000]$ & -- & -- & -- \\ 
        |\quad|\quad+Conv2d & $[\text{N},32,32,1000]$ & $[\text{N},32,32,1000]$ & $[1,7]$ & 1 & 7,168 \\ 
        |\quad|\quad+BatchNorm2d & $[\text{N},32,32,1000]$ & $[\text{N},32,32,1000]$ & -- & -- & 64 \\ 
        |\quad|\quad+ReLU & $[\text{N},32,32,1000]$ & $[\text{N},32,32,1000]$ & -- & -- & -- \\ 
        |\quad+Temporal ResBlock & $[\text{N},32,32,1000]$ & $[\text{N},32,32,1000]$ & -- & -- & 16,512 \\
        +Stage 2 & $[\text{N},32,32,1000]$ & $[\text{N},64,32,500]$ & -- & -- & -- \\ 
        |\quad+Temporal ResBlock (+ Downsampling) & $[\text{N},32,32,1000]$ & $[\text{N},64,32,500]$ & -- & -- & -- \\ 
        |\quad|\quad+Downsampling & $[\text{N},32,32,1000]$ & $[\text{N},64,32,500]$ & -- & -- & -- \\ 
        |\quad|\quad|\quad+Conv2d & $[\text{N},32,32,1000]$ & $[\text{N},64,32,500]$ & $[1,1]$ & 1 & 2,048 \\ 
        |\quad|\quad|\quad+BatchNorm2d & $[\text{N},64,32,500]$ & $[\text{N},64,32,500]$ & -- & -- & 128 \\ 
        |\quad|\quad+Conv2d & $[\text{N},32,32,1000]$ & $[\text{N},64,32,500]$ & $[1,7]$ & 1 & 14,336 \\ 
        |\quad|\quad+BatchNorm2d & $[\text{N},64,32,500]$ & $[\text{N},64,32,500]$ & -- & -- & 128 \\ 
        |\quad|\quad+ReLU & $[\text{N},64,32,500]$ & $[\text{N},64,32,500]$ & -- & -- & -- \\ 
        |\quad|\quad+Conv2d & $[\text{N},64,32,500]$ & $[\text{N},64,32,500]$ & $[1,5]$ & 1 & 20,480 \\ 
        |\quad|\quad+BatchNorm2d & $[\text{N},64,32,500]$ & $[\text{N},64,32,500]$ & -- & -- & 128 \\ 
        |\quad|\quad+ReLU & $[\text{N},64,32,500]$ & $[\text{N},64,32,500]$ & -- & -- & -- \\ 
        |\quad+Temporal ResBlock & $[\text{N},64,32,500]$ & $[\text{N},64,32,500]$ & -- & -- & 49,408 \\ 
        +Stage 3 & $[\text{N},64,32,500]$ & $[\text{N},128,32,250]$ & -- & -- & -- \\ 
        |\quad+Mixed ResBlock (+ Downsampling) & $[\text{N},64,32,500]$ & $[\text{N},128,32,250]$ & -- & -- & -- \\ 
        |\quad|\quad+Downsampling & $[\text{N},64,32,500]$ & $[\text{N},128,32,250]$ & -- & -- & -- \\ 
        |\quad|\quad|\quad+Conv2d & $[\text{N},64,32,500]$ & $[\text{N},128,32,250]$ & $[1,1]$ & 1 & 8,192 \\ 
        |\quad|\quad|\quad+BatchNorm2d & $[\text{N},128,32,250]$ & $[\text{N},128,32,250]$ & -- & -- & 256 \\ 
        |\quad|\quad+Conv2d & $[\text{N},64,32,500]$ & $[\text{N},128,32,250]$ & $[1,5]$ & 1 & 40,960 \\ 
        |\quad|\quad+BatchNorm2d & $[\text{N},128,32,250]$ & $[\text{N},128,32,250]$ & -- & -- & 256 \\ 
        |\quad|\quad+ReLU & $[\text{N},128,32,250]$ & $[\text{N},128,32,250]$ & -- & -- & -- \\ 
        |\quad|\quad+Conv2d & $[\text{N},128,32,250]$ & $[\text{N},128,32,250]$ & $[5,1]$ & 1 & 81,920 \\ 
        |\quad|\quad+BatchNorm2d & $[\text{N},128,32,250]$ & $[\text{N},128,32,250]$ & -- & -- & 256 \\ 
        |\quad|\quad+ReLU & $[\text{N},128,32,250]$ & $[\text{N},128,32,250]$ & -- & -- & -- \\ 
        |\quad+Mixed ResBlock & $[\text{N},128,32,250]$ & $[\text{N},128,32,250]$ & -- & -- & 164,352 \\ 
        +Stage 4 & $[\text{N},128,32,250]$ & $[\text{N},256,32,250]$ & -- & -- & -- \\ 
        |\quad+Spatial ResBlock (+ Downsampling) & $[\text{N},128,32,250]$ & $[\text{N},256,32,250]$ & -- & -- & -- \\ 
        |\quad|\quad+Downsampling & $[\text{N},128,32,250]$ & $[\text{N},256,32,250]$ & -- & -- & -- \\ 
        |\quad|\quad|\quad+Conv2d & $[\text{N},128,32,250]$ & $[\text{N},256,32,250]$ & $[1,1]$ & 1 & 32,768 \\ 
        |\quad|\quad|\quad+BatchNorm2d & $[\text{N},256,32,250]$ & $[\text{N},256,32,250]$ & -- & -- & 512 \\ 
        |\quad|\quad+Conv2d & $[\text{N},128,32,250]$ & $[\text{N},256,32,250]$ & $[5,1]$ & 1 & 163,840 \\ 
        |\quad|\quad+BatchNorm2d & $[\text{N},256,32,250]$ & $[\text{N},256,32,250]$ & -- & -- & 512 \\ 
        |\quad|\quad+ReLU & $[\text{N},256,32,250]$ & $[\text{N},256,32,250]$ & -- & -- & -- \\ 
        |\quad|\quad+Conv2d & $[\text{N},256,32,250]$ & $[\text{N},256,32,250]$ & $[3,1]$ & 1 & 196,608 \\ 
        |\quad|\quad+BatchNorm2d & $[\text{N},256,32,250]$ & $[\text{N},256,32,250]$ & -- & -- & 512 \\ 
        |\quad|\quad+ReLU & $[\text{N},256,32,250]$ & $[\text{N},256,32,250]$ & -- & -- & -- \\ 
        |\quad+Spatial ResBlock & $[\text{N},256,32,250]$ & $[\text{N},256,32,250]$ & -- & -- & -- \\ 
        +AdaptiveAvgPool2d & $[\text{N},256,32,250]$ & $[\text{N},256,1,1]$ & -- & -- & -- \\ 
        +Linear & $[\text{N},256]$ & $[\text{N},1]$ & -- & -- & 257 \\ 
        \bottomrule 
    \end{tabular}
    \label{tab:summaryTResNer}
\end{table*}
}

\texttt{
\begin{table*}
    \centering
    \caption{Summary Table of TransformEEG. Depthwise Conv Blocks include a skip connection.}
    \begin{tabular}{llllll}
        \toprule 
        Layer(type) & InputShape & OutputShape & KernelShape & Groups & Param\# \\ 
        \midrule 
        TransformEEG & $[\text{N},32,2000]$ & $[\text{N},1]$ & -- & -- & -- \\ 
        +Depthwise Tokenizer & $[\text{N},32,2000]$ & $[\text{N},128,498]$ & -- & -- & -- \\ 
        |\quad+Depthwise Conv Block & $[\text{N},32,2000]$ & $[\text{N},64,999]$ & -- & -- & -- \\ 
        |\quad|\quad+Conv1d & $[\text{N},32,2000]$ & $[\text{N},64,2000]$ & $[5]$ & 32 & 384 \\ 
        |\quad|\quad+BatchNorm1d & $[\text{N},64,2000]$ & $[\text{N},64,2000]$ & -- & -- & 128 \\ 
        |\quad|\quad+ELU & $[\text{N},64,2000]$ & $[\text{N},64,2000]$ & -- & -- & -- \\ 
        |\quad|\quad+AvgPool1d & $[\text{N},64,2000]$ & $[\text{N},64,999]$ & $[4]$ & -- & -- \\ 
        |\quad|\quad+Dropout1d & $[\text{N},64,999]$ & $[\text{N},64,999]$ & -- & -- & -- \\  
        |\quad|\quad+Conv1d & $[\text{N},64,999]$ & $[\text{N},64,999]$ & $[5]$ & 64 & 384 \\ 
        |\quad|\quad+BatchNorm1d & $[\text{N},64,999]$ & $[\text{N},64,999]$ & -- & -- & 128 \\ 
        |\quad|\quad+ELU & $[\text{N},64,999]$ & $[\text{N},64,999]$ & -- & -- & -- \\ 
        |\quad+Depthwise Conv Block & $[\text{N},64,999]$ & $[\text{N},128,498]$ & -- & -- & -- \\ 
        |\quad|\quad+Conv1d & $[\text{N},64,999]$ & $[\text{N},128,999]$ & $[5]$ & 64 & 768 \\ 
        |\quad|\quad+BatchNorm1d & $[\text{N},128,999]$ & $[\text{N},128,999]$ & -- & -- & 256 \\ 
        |\quad|\quad+ELU & $[\text{N},128,999]$ & $[\text{N},128,999]$ & -- & -- & -- \\ 
        |\quad|\quad+AvgPool1d & $[\text{N},128,999]$ & $[\text{N},128,498]$ & $[4]$ & -- & -- \\ 
        |\quad|\quad+Dropout1d & $[\text{N},128,498]$ & $[\text{N},128,498]$ & -- & -- & -- \\ 
        |\quad|\quad+Conv1d & $[\text{N},128,498]$ & $[\text{N},128,498]$ & $[5]$ & 128 & 768 \\ 
        |\quad|\quad+BatchNorm1d & $[\text{N},128,498]$ & $[\text{N},128,498]$ & -- & -- & 256 \\ 
        |\quad|\quad+ELU & $[\text{N},128,498]$ & $[\text{N},128,498]$ & -- & -- & -- \\ 
        +Transformer Encoder & $[\text{N},498,128]$ & $[\text{N},498,128]$ & -- & -- & -- \\
        |\quad+Permute & $[\text{N},498,128]$ & $[\text{N},128,498]$ & -- & -- & -- \\ 
        |\quad+TransformerEncoderLayer & $[\text{N},498,128]$ & $[\text{N},498,128]$ & -- & -- & -- \\ 
        |\quad|\quad+MultiheadAttention & $[\text{N},498,128]$ & $[\text{N},498,128]$ & -- & -- & 66,048 \\ 
        |\quad|\quad+Dropout & $[\text{N},498,128]$ & $[\text{N},498,128]$ & -- & -- & -- \\ 
        |\quad|\quad+LayerNorm & $[\text{N},498,128]$ & $[\text{N},498,128]$ & -- & -- & 256 \\ 
        |\quad|\quad+Linear & $[\text{N},498,128]$ & $[\text{N},498,128]$ & -- & -- & 16,512 \\ 
        |\quad|\quad+Dropout & $[\text{N},498,128]$ & $[\text{N},498,128]$ & -- & -- & -- \\ 
        |\quad|\quad+Linear & $[\text{N},498,128]$ & $[\text{N},498,128]$ & -- & -- & 16,512 \\ 
        |\quad|\quad+Dropout & $[\text{N},498,128]$ & $[\text{N},498,128]$ & -- & -- & -- \\ 
        |\quad|\quad+LayerNorm & $[\text{N},498,128]$ & $[\text{N},498,128]$ & -- & -- & 256 \\ 
        |\quad+TransformerEncoderLayer & $[\text{N},498,128]$ & $[\text{N},498,128]$ & -- & -- & 99,584 \\
        |\quad+Permute & $[\text{N},498,128]$ & $[\text{N},128,498]$ & -- & -- & -- \\
        +Classification MLP & $[\text{N},128,498]$ & $[\text{N},1]$ & -- & -- & -- \\
        |\quad+AdaptiveAvgPool1d & $[\text{N},128,498]$ & $[\text{N},128,1]$ & -- & -- & -- \\ 
        |\quad+Linear & $[\text{N},128]$ & $[\text{N},64]$ & -- & -- & 8,256 \\ 
        |\quad+LeakyReLU & $[\text{N},64]$ & $[\text{N},64]$ & -- & -- & -- \\ 
        |\quad+Linear & $[\text{N},64]$ & $[\text{N},1]$ & -- & -- & 65 \\ 
        \bottomrule  
    \end{tabular}
    \label{tab:summarytransformeeg}
\end{table*}
}

\clearpage
\textcolor{white}{.}